\documentclass[pdflatex,sn-mathphys-num]{sn-jnl}

\usepackage{graphicx}%
\usepackage{multirow}%
\usepackage{amsmath,amssymb,amsfonts}%
\usepackage{amsthm}%
\usepackage{mathrsfs}%
\usepackage[title]{appendix}%
\usepackage{xcolor}%
\usepackage{textcomp}%
\usepackage{manyfoot}%
\usepackage{booktabs}%
\usepackage{algorithm}%
\usepackage{algorithmicx}%
\usepackage{algpseudocode}%
\usepackage{listings}%

\usepackage{makecell}
\usepackage{booktabs}

\usepackage{amssymb}
\usepackage{pifont}

\usepackage[T1]{fontenc}
\usepackage{lmodern}

\theoremstyle{thmstyleone}%
\theoremstyle{thmstyletwo}%

\theoremstyle{thmstylethree}%

\begin{document}

\title[Article Title]{Blurring Modal Boundaries: A Unified Survey from Single- to Multi-Modal Person Re-ldentification}
\author{Xiao Wang$^{1,2}$, Bing Wang$^{1,2}$, Bin Yang$^{3,}$\thanks{Corresponding author.}, Cuiqun Chen$^4$, Xin Xu$^{1,2}$, Mang Ye$^{3}$\thanks{Corresponding author.}}

\email{yangbin\_cv@whu.edu.cn;yemang@whu.edu.cn}



\affil[1]{\orgdiv{School of Computer Science and Technology}, \orgname{Wuhan University of Science and Technology}, \orgaddress{\street{Wuhan}, \city{Hubei}, \postcode{430065}, \country{China}}}

\affil[2]{\orgdiv{Hubei Province Key Laboratory of Intelligent Information Processing and Real-time Industrial System}, \orgname{Wuhan University of Science and Technology}, \orgaddress{\street{Wuhan}, \city{Hubei}, \postcode{430065}, \country{China}}}

\affil[3]{\orgdiv{School of Computer}, \orgname{Wuhan University}, \orgaddress{\street{Wuhan}, \city{Hubei}, \postcode{430072}, \country{China}}}

\affil[4]{\orgdiv{School of Computer}, \orgname{Anhui University}, \orgaddress{\street{Hefei}, \city{Anhui}, \postcode{232002}, \country{China}}}

\abstract{Person re-identification (ReID) serves as a critical component in intelligent surveillance systems, aiming to match identities across disjoint camera networks. While traditional methods primarily rely on single-modal RGB imagery, they are often constrained by environmental challenges such as low illumination and occlusion. To overcome these limitations, the field is rapidly evolving toward cross-modal and multi-modal paradigms. This survey presents a comprehensive overview of this transition, systematically reviewing key cross-modal tasks including visible-infrared (VI-ReID), text-image (TI-ReID), sketch-based (Sketch-ReID), and the emerging Non-Line-of-Sight (NLOS) ReID, which extends perception beyond direct visibility. Furthermore, we examine tri-spectral and multi-modal fusion ReID, discussing how complementary information from diverse sensors enhances robustness. Beyond summarizing datasets, challenges, and methodologies, we propose a Transformer-based baseline framework for visible-infrared ReID, designed to effectively capture modality-invariant features. Finally, based on the current landscape, we outline several promising directions for future research.}



\maketitle

\section{Introduction}\label{sec1}

Person re-identification (ReID) \cite{CaoFXGWZ26,DaiSLTD25,ZhangTJF25} is a fundamental computer vision task that aims to recognize the same individual across non-overlapping camera views. The core scientific challenge lies in achieving accurate and robust cross-view matching under complex conditions such as background clutter, pose variations, occlusions, and illumination changes~\cite{zheng2016person}. As a core component of intelligent surveillance systems, ReID facilitates the efficient retrieval of persons of interest from massive-scale video streams, thereby providing essential technical support for critical public safety applications such as urban security, traffic management, and societal governance.
Early studies on person ReID primarily focused on the supervised learning paradigm within the visible spectrum, where high-quality RGB images with explicit identity labels were utilized to achieve cross-camera pedestrian matching through deep feature extraction and metric learning. Such approaches \cite{wieczorek2021unreasonable, wang2019spatial, alnissany2023modified, zang2021learning, shi2023boosting, gong2024crossmodality} have achieved remarkable success on standardized datasets such as Market-1501 \cite{market1501}. 
However, such approaches suffer from substantial limitations when applied to complex real-world scenarios. In low-light environments or adverse weather conditions, insufficient reflection of visible light frequently leads to severe degradation in image quality, manifesting as blurring, color distortion, and the loss of contour information. These issues consequently compromise the stability and reliability of identity feature extraction.

\begin{figure}
    \centering
    \includegraphics[width=1\linewidth]{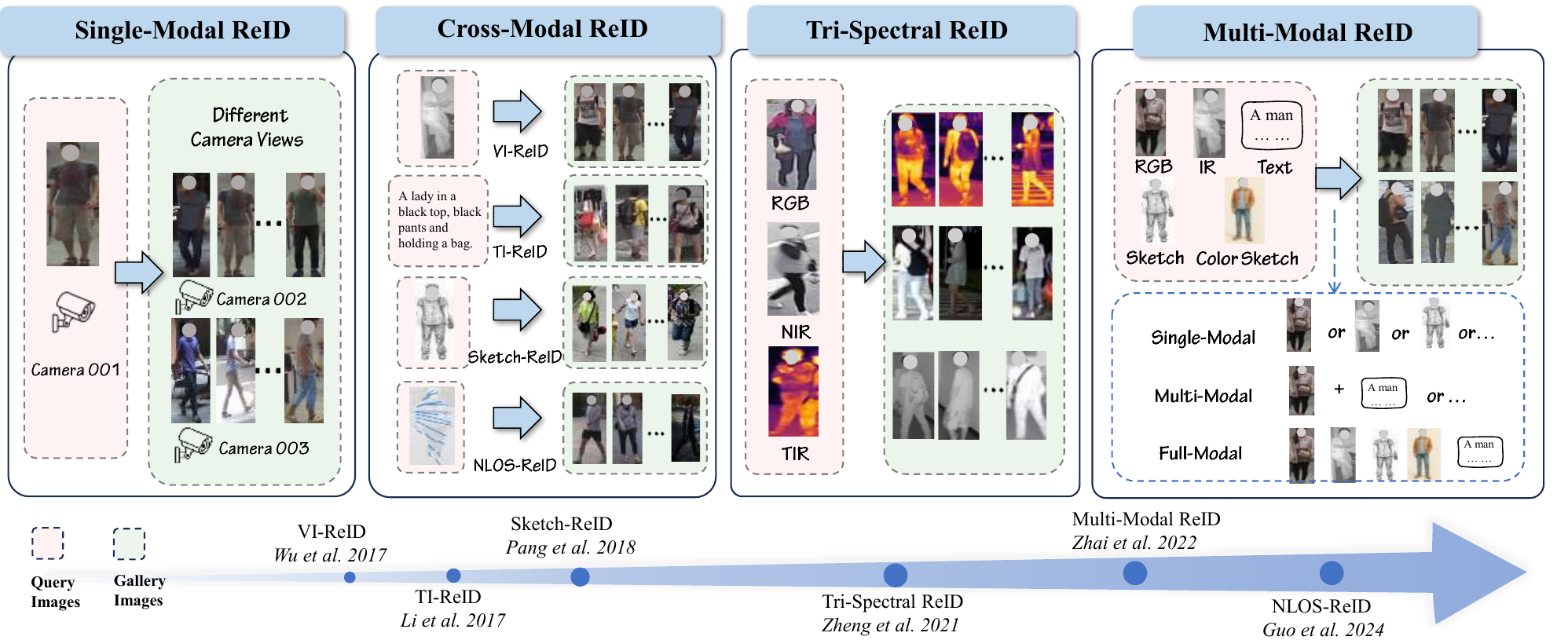}
    \caption{This diagram classifies person re-identification approaches into four categories: Single-Modal ReID, Cross-Modal ReID, Tri-Spectral ReID, and Multi-Modal ReID. It illustrates the evolution from basic single-modal systems to advanced multi-modal frameworks that integrate various data sources, thereby enhancing identification robustness and accuracy. This progression reflects significant advancements in the effectiveness of person re-identification methodologies.
}
    \label{fig:introduction}
    \vspace{-4mm}
\end{figure}

In response to the aforementioned limitations, researchers have progressively extended person ReID from single-modal recognition to cross-modal and, subsequently, tri-spectral and multi-modal settings. This evolution reflects the increasing emphasis on exploiting heterogeneous information to address more realistic and complex retrieval scenarios. Fig.~\ref{fig:introduction} summarizes this developmental trajectory, illustrating the progression of person ReID from single-modal recognition to cross-modal, tri-spectral, and multi-modal paradigms.

To provide a clearer understanding of this evolution, we organize existing
  person re-identification tasks according to their retrieval protocols and primary learning objectives, rather than merely the number of involved modalities. Specifically, cross-modal person re-identification performs identity retrieval between heterogeneous query and gallery domains. Its primary learning
  objective is heterogeneous modality alignment, which aims to reduce the modality gap while preserving identity consistency across different representations. Representative tasks include visible--infrared ReID (VI-ReID), text--image ReID
  (TI-ReID), and sketch--image ReID (Sketch-ReID). We also discuss NLOS-related ReID, particularly emerging signal-to-visual settings that associate wireless, radar, or sparse geometric observations with visual identity representations.
  These settings share the challenge of heterogeneous alignment.

  Meanwhile, tri-spectral person re-identification has gradually developed into a dedicated research direction. Its distinction is not determined solely by the use of three sensing modalities. Existing tri-spectral methods place particular
  emphasis on spectral-aware representation learning by modeling the spectral characteristics, inter-spectral consistency, and complementary identity evidence across RGB, near-infrared (NIR), and thermal-infrared (TIR) observations.
  Accordingly, we review tri-spectral ReID in a dedicated section to summarize its methodological characteristics, technical challenges, and open research questions. 

\begin{figure}
      \centering
      \includegraphics[width=1\linewidth]{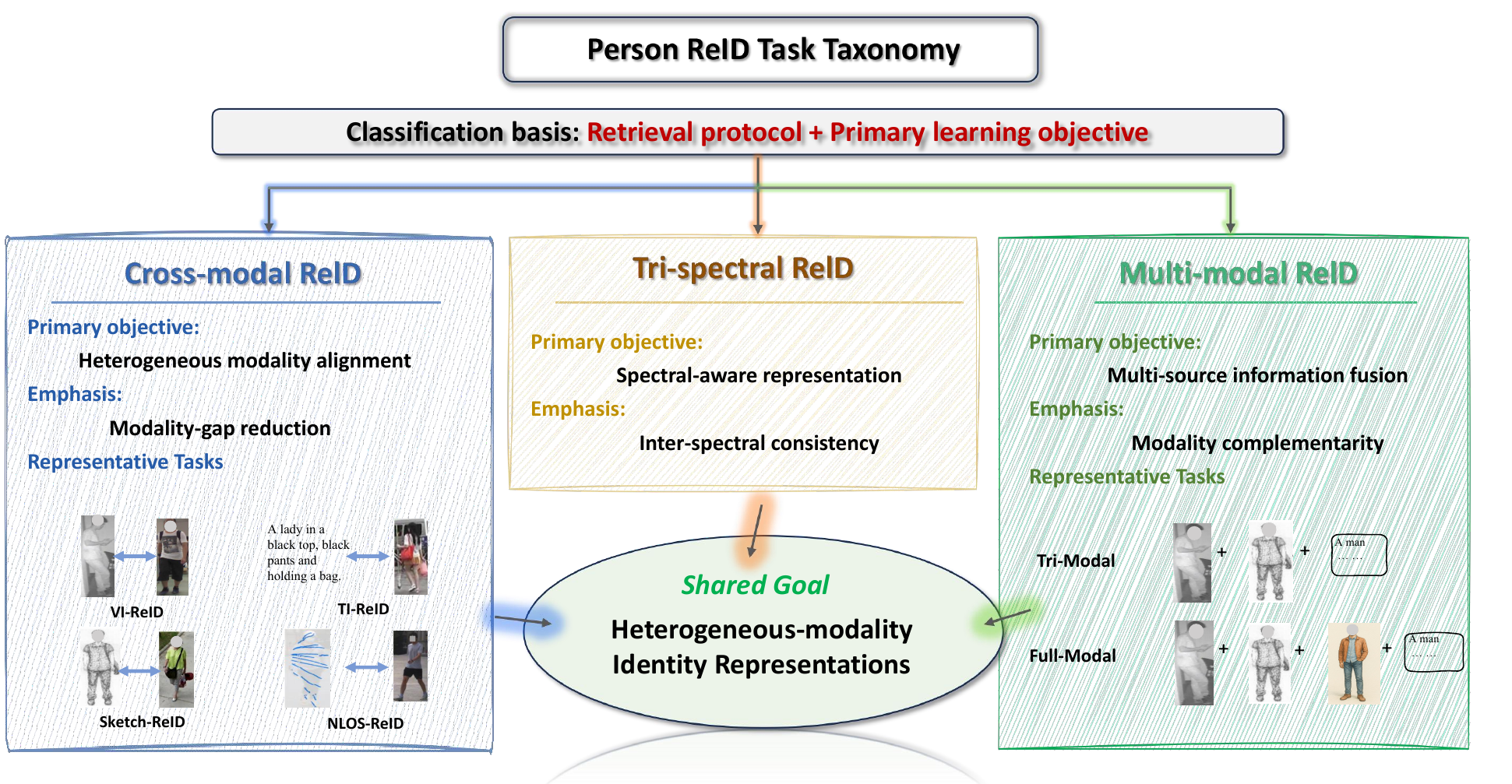}
      \caption{Proposed taxonomy of person re-identification tasks. Existing ReID tasks are systematically categorized based on the retrieval protocol and primary learning objective into cross-modal, tri-spectral, and multi-modal ReID. Despite their different modality settings and optimization objectives, all aim to learn robust identity representations across heterogeneous modalities.}
      \label{fig:taxonomy}
  \end{figure}

  In contrast, multi-modal person re-identification jointly exploits multiple available information sources or flexible modality combinations for identity representation and retrieval. Its primary learning objective is multi-source
  information fusion, with an emphasis on exploiting modality complementarity while controlling cross-source redundancy. Representative settings include joint fusion of RGB and infrared observations, as well as more flexible frameworks that combine RGB, infrared, text, sketch, and color-pencil representations. Cross-modal, tri-spectral, and multi-modal ReID therefore represent different paradigms for exploiting heterogeneous modalities, distinguished primarily by their retrieval protocol and learning objectives while sharing the common goal of robust cross-modal identity representation. Based on these perspectives, Fig. \ref{fig:taxonomy} illustrates the taxonomy adopted throughout this survey.



The Re-ID task is increasingly responsive to complex perception demands, real-world deployment challenges, and multi-modal integration. Several surveys have focused on Re-ID and cross-modal Re-ID. D’Orazio and Cicirelli \cite{d2012people} provide an early systematic survey on person re-identification and tracking across non-overlapping camera views, emphasizing appearance-based techniques. Bedagkar et al. \cite{bedagkar2014survey} categorize person re-identification into Contextual and Non-contextual Methods based on visual features. Satta et al. \cite{satta2013appearance} offer a comprehensive review of methodologies for appearance descriptor construction. Zheng et al. \cite{zheng2016person} trace the evolution of person re-identification, differentiating image-based from video-based approaches and reviewing both handcrafted and deep learning methods. Leng et al. \cite{leng2019survey} present the first survey of open-world Re-ID, aligning closely with real-world applications. Ye et al. \cite{AGW} categorize Re-ID systems into open-world and closed-world settings and survey closed-world Re-ID through deep feature representation, metric learning, and ranking optimization. Zheng et al. \cite{zheng2022visible} provide an overview of VI-ReID research, comparing traditional Re-ID and VI-ReID. Chang et al. \cite{chang2024comprehensive} categorize VI-ReID studies based on application scenarios. Nguyen et al. \cite{nguyen2024tackling} delivers a review of domain shift challenges in three settings: Unsupervised Domain Adaptation ReID, Domain Generalizable ReID, and Lifelong ReID. Chen et al. \cite{chen2024person} analyze recent Re-ID research through specific application scenarios, while Wang et al. \cite{wang2019beyond} systematically classify heterogeneous person re-identification across various cross-modal contexts. Huang et al. \cite{huang2023deep} survey VI-ReID from four technical perspectives: modality-shared feature learning, modality-specific compensation, auxiliary information utilization, and data augmentation.

Through an examination of existing surveys, we find that existing reviews lack a comprehensive research landscape that spans diverse cross-modal and multi-modal tasks. In response, this paper presents a systematic overview covering three representative cross-modal tasks: VI-ReID, TI-ReID, Sketch-ReID and NLOS ReID, as well as tri-spectral scenarios (e.g., RGB-NIR-TIR) and multi-modal ReID. We analyze the technical evolution and key challenges across these settings, aiming to construct a more holistic research map that offers insights and guidance for future studies.
We present in Table \ref{tab:surveys} a comparison between our survey and existing surveys, highlighting the differences in covered tasks, modalities, and research perspectives.
In this work, we also establish a transformer-based baseline for VI-ReID. This baseline serves as a simple yet effective reference framework, offering insights into how Transformer architectures can be adapted to cross-modal person re-identification.

The major contributions of this article are outlined as follows: 

\begin{itemize}
    \item \textbf{Comprehensive coverage of diverse scenarios.} To the best of our knowledge, this is the first survey that systematically reviews cross-modal ReID across multiple scenarios, including visible–infrared , text-image , sketch-image and Non-Line-of-Sight settings, as well as multi-spectral and multi-modal extensions.
    \item \textbf{Detailed analysis of representative methods.} We provide an in-depth review of related studies in each subfield, discussing representative methods and datasets, thereby offering a structured understanding of the research landscape.
    \item \textbf{Introduction of a Transformer-based baseline.} We establish a simple yet effective Transformer-based framework tailored for VI-ReID, which serves as a reference model and highlights the potential of Transformer architectures in addressing cross-modal person re-identification.
\end{itemize}

\newcommand{\cmark}{\ding{51}} 
\newcommand{\xmark}{\ding{55}} 
\newcommand{\dash}{--}
\begin{table*}[htbp]
\centering
\setlength{\tabcolsep}{3pt} 
\footnotesize
\caption{Overview of survey papers on person re-identification and the modality scenarios they address. The covered modalities include single-modal, cross-modal (visible-infrared, text-image, visible-sketch and NLOS), tri-spectral, and multi-modal settings. A check mark "\cmark" indicates that the corresponding modality is discussed in the survey, while a cross mark “\xmark” denotes that the modality is not covered.}
\centering
  {\fontsize{7pt}{10pt}\selectfont 
\begin{tabular}{ll|c|cccc|c|c}
\toprule
\multirow{2}{*}{Surveys} & \multirow{2}{*}{Venue} & \multirow{2}{*}{Single-modal} & \multicolumn{4}{c|}{Cross-modal} & \multirow{2}{*}{Tri-spectral}& \multirow{2}{*}{Multi-modal} \\
 &  &  & V-I & T-I & V-S & NLOS &  & \\
\hline
D'Orazio et al.\cite{d2012people}& ICIP-12 & \cmark & \xmark & \xmark & \xmark & \xmark & \xmark & \xmark  \\
Satta et al.\cite{satta2013appearance}& arXiv-13 & \cmark & \xmark & \xmark & \xmark & \xmark & \xmark & \xmark  \\
Bedagkar-Gala et al.\cite{bedagkar2014survey}& IVC-14 & \cmark & \xmark & \xmark & \xmark & \xmark & \xmark & \xmark  \\
Gong et al.\cite{gong2014person}& VAB-14 & \cmark & \xmark & \xmark & \xmark & \xmark & \xmark & \xmark  \\
Zheng et al.\cite{zheng2016person}& arXiv-16 & \cmark & \xmark & \xmark & \xmark & \xmark & \xmark & \xmark  \\
Leng et al.\cite{leng2019survey} & TCSVT-19 & \cmark & \xmark & \xmark & \xmark & \xmark & \xmark & \xmark \\
Wang et al.\cite{wang2019beyond} & arXiv-19 & \xmark & \cmark & \cmark & \cmark & \xmark & \xmark & \xmark  \\
Ye et al.\cite{AGW} & TPAMI-22 & \cmark & \xmark & \xmark & \xmark & \xmark & \xmark & \xmark  \\
Zheng et al.\cite{zheng2022visible} & Electronics-22 & \xmark & \cmark & \xmark & \xmark & \xmark & \xmark & \xmark\\
Chang et al.\cite{chang2024comprehensive} & MTAP-24 & \xmark & \cmark & \xmark & \xmark & \xmark & \xmark & \xmark \\
Huang et al.\cite{huang2023deep} & INFFUS-23 & \xmark & \cmark & \xmark & \xmark & \xmark & \xmark & \xmark  \\
Nguyen et al.\cite{nguyen2024tackling} & CVPR-24 & \xmark & \cmark & \xmark & \xmark & \xmark & \xmark & \xmark  \\
Chen et al.\cite{chen2024person} & Mathematics-24 & \xmark & \cmark & \xmark & \xmark & \xmark & \xmark & \xmark \\
Ye et al.\cite{ye2025transformer} & IJCV-25 & \cmark & \cmark & \cmark & \cmark & \xmark & \xmark & \cmark \\
\hline
\textbf{Ours} & - & \xmark & \cmark & \cmark & \cmark & \cmark & \cmark & \cmark \\
\bottomrule
\end{tabular}
}
\label{tab:surveys}
\end{table*}

\section{Cross-modal Person Re-Identification}


In contrast to conventional Re-Identification (ReID) methods that operate within a single modality,
cross-modal person re-identification performs identity retrieval between heterogeneous query and
gallery domains. Its central objective is to reduce the representation discrepancy induced by different
sensing or descriptive modalities while preserving identity-discriminative information. This section systematically reviews these representative cross-modal ReID tasks, focusing on task definitions, dataset construction, key challenges, and notable methodologies.
\subsection{Visible-Infrared ReID}


 Visible-Infrared person re-identification (VI-ReID) is a cross-modal task that aims to identify the same individual by matching images captured by visible and infrared cameras~\cite{Wu_2017_ICCV}. This task expands the scope of traditional single-modal person re-identification (ReID) by tackling the significant performance degradation that occurs under challenging lighting conditions, such as nighttime or low-illumination scenarios. The core challenge of visible-infrared person re-identification (VI-ReID) resides in mitigating the feature distribution shift induced by distinct imaging mechanisms: visible images deliver abundant color and texture information but are highly susceptible to degradation under low-light environments, while infrared images, which rely on thermal radiation, exhibit robust imaging performance under adverse illumination conditions yet lack fine-grained appearance details. Depending on whether the training process leverages cross-modal matching pairs and identity annotations, existing research on VI-ReID can be generally classified into supervised and unsupervised paradigms. In this section, we summarize the key challenges confronted by each paradigm and conduct a comprehensive review of representative methodologies.


\subsubsection{Datasets}
Public benchmark datasets specifically designed for VI-ReID have played a crucial role in advancing this field.  The details of the datasets are summarized in Table \ref{tab:datasets-VIReID}.

SYSU-MM01~\cite{Wu_2017_ICCV} is the first large-scale VI-ReID dataset, containing 491 identities with a total of 287,628 visible images and 15,792 infrared images captured by six cameras (labeled 1 to 6) across diverse scenes. It supports two evaluation modes: full search and indoor search. In full search mode, visible images from cameras 1, 2, 4, and 5 form the gallery, while infrared images from cameras 3 and 6 are used as the query set. In indoor search mode, only visible images from cameras 1 and 2 are included in the gallery, with the query set remaining unchanged. 

RegDB~\cite{regdb} contains 412 identities, with each individual associated with 10 visible images and 10 infrared images captured in real-world daytime and nighttime environments. It includes two evaluation modes: Visible To Infrared, where visible images serve as the query set and infrared images as the gallery, and Infrared To Visible, where the roles are reversed.


LLCM (Large-scale Low-light Condition Multi-modal) \cite{DEEN} dataset is the first large scale VI-ReID benchmark specifically constructed for nighttime low-light scenarios commonly encountered in real-world surveillance. It is collected using a 9-camera network deployed in low-light environments and contains 46,767 images of 1,064 identities, with 30,921 images from 713 identities in the training set (16,946 VIS and 13,975 IR) and 13,909 images from 351 identities in the testing set (8,680 VIS and 7,166 IR). The dataset covers both daytime and nighttime conditions, various climates, and diverse clothing styles, while also including real-world challenges such as motion blur, pose variation, occlusion, and low resolution. Similar to the RegDB dataset, two evaluation modes are adopted, i.e., visible to infrared and infrared to visible. 


\begin{table*}
  \caption{Statistics of VI-ReID related datasets, including the number of identities, images, cameras, and the presence of low-light conditions.}
  \label{tab:datasets-VIReID}
  \centering
  {\fontsize{8pt}{12pt}\selectfont 
  \begin{tabular}{ccccc}
    \toprule
    Datasets & Identitiesr & Image number & Camera number & Low-light \\
    \midrule
    SYSU-MM01 \cite{Wu_2017_ICCV} & 491 & 303,420 & 6 & \xmark \\
    RegDB \cite{regdb}  & 412 & 8,240 & 2 & \xmark \\
    LLCM \cite{DEEN}   & 1,064 & 46,767 & 18 & \cmark \\
    \bottomrule
  \end{tabular}}
\end{table*}

\textbf{Summary:} These benchmarks provide complementary evaluation conditions rather than a single unified measure of VI-ReID
  robustness. SYSU-MM01 emphasizes scale and camera diversity, RegDB offers a more balanced and controlled bidirectional
  setting, and LLCM focuses on challenging low-light surveillance. To obtain a more reliable assessment of cross-
  environment generalization, future studies should report cross-dataset evaluation, camera- or environment-disjoint
  testing, and robustness under degraded or unavailable modalities, rather than relying solely on within-dataset Rank-1
  accuracy and mAP.


\subsubsection{Supervised Visible-Infrared Person ReID} 
\begin{figure}
    \centering
    \includegraphics[width=1\linewidth]{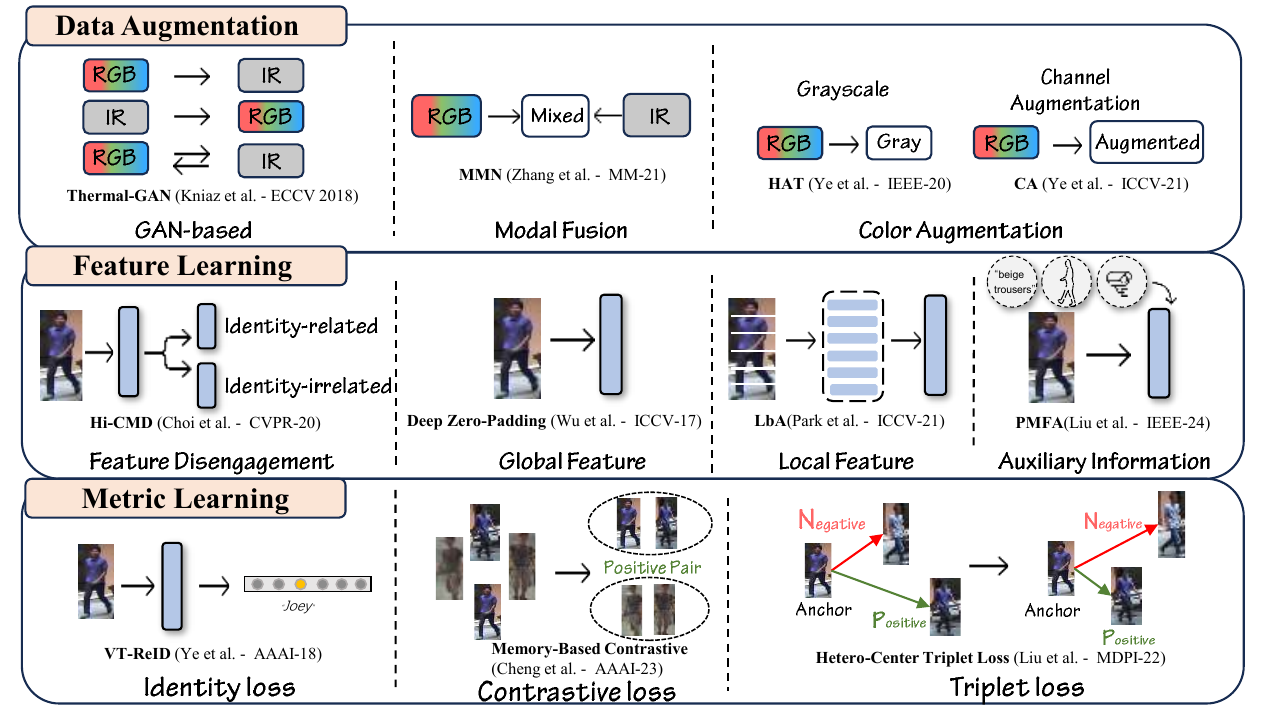}
    \caption{A taxonomy of supervised VI-ReID methods, categorized into data augmentation, feature learning, and metric learning. Representative works are listed under each sub-direction.}

    \label{fig:sup}
\end{figure}

Supervised Visible-Infrared Person Re-ID aims to develop discriminative, modality-invariant representations from labeled cross-modal identity annotations, facilitating accurate identity matching across diverse modalities. To tackle challenges encountered during various training stages, existing research can be classified into three main directions: data augmentation, feature learning, and metric learning. This section systematically reviews prominent techniques within each direction, with an overview of major approaches and representative studies presented in Fig. \ref{fig:sup}, and tabulate a comprehensive performance comparison of representative methods in Table \ref{tab:sup}.

\textbf{Data Augmentation for cross-modal Robustness:}
In supervised VI-ReID, substantial modality discrepancies, limited training samples, and complex environmental conditions are major factors constraining model performance and generalization. To mitigate these challenges, data augmentation has been widely employed as an image-level optimization strategy. By expanding the sample distribution, increasing image diversity, and simulating complex scenarios, data augmentation enhances model robustness to modality variations and facilitates more effective cross-modal feature alignment and discrimination.
Early augmentation approaches in VI-ReID primarily relied on basic image-level perturbations, such as random horizontal flipping, cropping, and erasing \cite{zhong2020random}, aiming to improve generalization under limited training data. However, as the demand for higher cross-modal alignment accuracy and more discriminative feature learning has grown, these conventional augmentation strategies have proven insufficient. Consequently, researchers have proposed more targeted cross-modal augmentation methods to address these limitations.

Inspired by single-modal ReID, some studies employ generative adversarial network (GAN)-based methods to map visible and infrared images into a unified modality domain, thereby simplifying cross-modal matching into a single-modal problem. Kniaz et al. \cite{kniaz2018thermalgan} developed Thermal-GAN, a color-to-thermal re-identification framework that models thermal appearances conditioned on visible inputs. Despite its effectiveness, direct visible-to-thermal translation loses critical channel cues, including textures and color distributions essential for discriminative learning. Zhong et al. \cite{zhong2020visible} proposed a reverse modeling approach that generates pseudo-color images from infrared inputs (i.e., image colorization), making semantic and texture details in infrared data explicitly visible in the color space. Wang et al. \cite{wang2019learning} employed dual GANs for bidirectional mapping between visible and infrared modalities, embedding the translated results into a shared feature space to support cross-modal matching. Zhu et al. \cite{zhu2017unpaired} synthesized new modality samples and jointly trained models with both synthetic and real images to enhance alignment. Qi et al. \cite{qi2023generative} further incorporated paired information from the complementary modality and introduced a part-based cross-modal contrastive loss to provide consistent supervision for modality transition.
Conversely, Qian et al. \cite{qian2024pose} presented a pose-guided augmentation strategy that leverages attention mechanisms to generate pose-diverse, identity-consistent image pairs, thereby enriching the structural diversity of training data.

In contrast to methods that focus solely on modality translation, another research direction investigates the fusion of visual information from both visible and infrared modalities at the image level to create training samples with mixed-modality characteristics. This approach allows models to concurrently capture discriminative cues from both modalities during training. For example, Zhang et al. \cite{zhang2021towards} develop a nonlinear intermediate-modality generator to synthesize images with unified intermediate features, which are employed alongside original images for training. Huang et al. \cite{huang2022modality} horizontally divide RGB and IR images into multiple local regions and apply linear interpolation within each region using a modality-adaptive mixing strategy. Qian et al. \cite{qian2025visible} partition visible and infrared images into small patches and reassemble them to create novel training samples that integrate both modalities.

\begin{table*}[ht]
\centering
\footnotesize
\setlength{\tabcolsep}{7pt}
\caption{Performance comparison of supervised VI-ReID methods on SYSU-MM01, RegDB, and LLCM datasets in terms of Rank-1 accuracy and mAP.}
\resizebox{\textwidth}{!}{
\centering
  {\fontsize{12pt}{12pt}\selectfont 
\begin{tabular}{l|l|cc|cc|cc|cc|cc|cc}
\toprule
\multirow{3}{*}{} & \multirow{3}{*}{} & \multicolumn{4}{c|}{SYSU-MM01}  & \multicolumn{4}{c}{RegDB}  & \multicolumn{4}{c}{LLCM} \\
\cmidrule{3-14}
\multirow{1}{*}{Methods} & \multirow{1}{*}{Venue} & \multicolumn{2}{c|}{All Search} & \multicolumn{2}{c|}{Indoor Search} & \multicolumn{2}{c|}{VIS to IR} & \multicolumn{2}{c|}{IR to VIS} & \multicolumn{2}{c|}{VIS to IR} & \multicolumn{2}{c}{IR to VIS}  \\
  & & R-1  & mAP & R-1  & mAP & R-1  & mAP & R-1  & mAP & R-1  & mAP & R-1  & mAP \\
\hline
BDTR \cite{BDTR}& IJCAI-18& 17.0  & 19.7 & - &  - & 33.6 & 32.8 & 32.9  & 32.0 & - & - & - & -\\
D$^2$RL \cite{wang2019learning}& CVPR-19 & 28.9 & 29.2 & -  & - & 43.4  & 44.1 & -  & - & - & - & - & -\\
Hi-CMD \cite{choi2020hi}& CVPR-20& 34.9  & 35.9 & -  & - & 70.9  & 66.0 & -  & - & - & - & - & -\\
JSIA-ReID \cite{JSIA-ReID}& AAAI-20& 38.1 & 36.9 & 43.8  & 52.9 & 48.1 & 48.9 & 48.5  & 49.3 & - & - & - & -\\
AlignGAN \cite{AlignGAN}& ICCV-19& 42.4  & 40.7 & 45.9  & 54.3 & 57.9  & 53.6 & 56.3 & 53.4 & - & - & - & -\\
X-Modality \cite{X-modality}&AAAI-20& 49.9  & 50.7 & - & - & 62.2 & 60.2 & - & - & - & - & - & -\\
DDAG \cite{ye2020dynamic}& ECCV-20& 54.8  & 53.0 & 61.0 & 68.0 & 69.3 & 63.5  & 68.1 & 61.8 & 40.3 & 48.4 & 48.0 & 52.3 \\
AGW \cite{AGW}&TPAMI-21& 47.5  & 47.65 & 54.17 & 62.97 & 70.05 & 66.37 & 70.49 & 65.90 & 43.6 & 51.8 & 51.5 & 55.3 \\
LbA \cite{park2021learning}&ICCV-21& 55.4  & 53.0 & 61.0 & 68.0 & 69.3 & 63.5 & 68.1 & 61.8 & 43.8  & 53.1 & 50.8 & 55.9 \\
NFS \cite{NFS}&CVPR-21& 56.9 & 55.5 & 62.8 & 69.8 & 80.5 & 72.1 & 78.0 & 69.8 & - & - & - & -\\
CM-NAS \cite{CM-NAs}&ICCV-21& 60.8 & 58.9 & 68.0 & 52.4 & 82.8 & 79.3 & 81.7 & 77.6 & - & - & - & -\\
MCLNet \cite{hao2021cross}&ICCV-21& 65.4 & 62.0 & 72.6 & 76.6 & 80.3 & 73.1 & 75.9 & 69.5 & - & - & - & -\\
FMCNet \cite{zhang2022fmcnet}& CVPR-22& 66.3 & 62.5 & 68.2 & 74.1 & 89.1 & 84.4 & 88.4 & 83.9 & - & - & - & -\\
SMCL \cite{wei2021syncretic}& ICCV-21& 67.4 & 61.8 & 68.8 & 75.6 & 83.9 & 79.8 & 83.1 & 78.6 & - & - & - & -\\
DART \cite{yang2022learning}&CVPR-22& 68.7 & 66.3 & 72.5 & 78.2 & 83.6 & 75.7 & 82.0 & 73.8 & 52.2 & 59.8 & 60.4 & 63.2 \\
CAJ \cite{yeCA}&ICCV-21& 69.9 & 66.9 & 76.2 & 79.6 & 91.6 & 84.1 & 87.5 & 80.5 & 48.8 & 56.6 & 56.5 & 59.8 \\
MPANet \cite{MPANet}&CVPR-21& 70.6  & 68.2 & 76.7 & 81.0 & 86.3  & 74.3 & 85.8 & 70.6 & - & - & - & -\\
MMN \cite{zhang2021towards}& MM-21& 70.6 & 66.9 & 76.2 & 79.6 & 91.6 & 84.1 & 87.5 & 80.5 & 52.5 & 58.9 & 59.9 & 62.7\\
DCLNet \cite{DCLNet}& MM-22& 70.8 & 65.3 & 73.5 & 76.8 & 81.2 & 74.3 & 78.0 & 70.6 & - & - & - & -\\
MAUM \cite{MAUM}& CVPR-22& 71.7 & 68.8 & 77.0 & 81.9 & 87.9 & 85.1 & 87.0 & 84.3 & - & - & - & -\\
DEEN  \cite{DEEN}& CVPR-23& 74.7 & 71.8 & 80.3 & 83.3 & 91.1 & 85.1 & 89.5 & 83.4 & 54.9 & 62.9 & 62.5 & 65.8\\
DEN \cite{DEN}& WACV-24 & 76.36 & 71.3 & 83.56 & 84.65 & 95.34 & 90.21 &94.98 & 90.24 & - & - & - & - \\

PartMix\cite{PartMix} & CVPR-23 & 77.78 & 74.62 & 81.52 & 84.38 & - & - &- &- & - & - & - & - \\
PMCM \cite{qian2025visible} & PR-25 & 75.5& 71.2 & 81.5 & 84.3 & 93.1 & 89.6 & 91.4 & 87.2 & - & - & - & -  \\
\bottomrule
\end{tabular}}}
\label{tab:sup}
\end{table*}

GAN-based generative methods and image-level mixing strategies, though demonstrating effectiveness in alleviating modality discrepancies, both exhibit inherent limitations. The former often introduces artifacts, distortions, and uncontrollable noise, whereas the latter may struggle to preserve fine-grained structural consistency. These shortcomings can ultimately undermine the reliability of model learning.

Recent studies have explored more controllable and stable augmentation techniques. Among them, channel-level augmentation directly manipulates image color channels or spectral responses to suppress modality-specific color cues and guide the model to focus on more discriminative structural features. This approach offers a more robust solution for modality alignment while avoiding the artifacts commonly introduced by generative models.

In earlier work, Ye et al. \cite{ye2020visible} design the Homogeneous Augmented Tri-modal (HAT) framework, which converts visible images into grayscale images, introduces intermediate modality images combining visible and infrared styles, and jointly trains the three modalities to reduce modality discrepancy. Liu et al. \cite{liu2021sfanet} extend the grayscale idea but simplify the processing pipeline by directly replacing the original visible image with its grayscale counterpart for model training, thereby preserving structural features while weakening modality gaps. Zhang et al. \cite{zhang2024multi} propose a staged strategy: in the first stage, auxiliary modality pre-training is conducted using grayscale-balanced images; in the second stage, original images are reintroduced to mitigate sensitivity to color features during model training. Subsequently, Ye et al. \cite{ye2023channel} propose an effective random channel exchange strategy, which disrupts the original color-channel relationships and reduces modality-specific correlations. Due to its simplicity and effectiveness, this method becomes a common augmentation technique in cross-modal ReID, as illustrated in the Fig. \ref{fig:sup}.

In addition, some studies explore various innovative approaches for channel-level data augmentation. Zhang et al. \cite{zhang2022dual} map single-channel infrared images into pseudo-RGB three-channel format to achieve channel-wise dual semantic alignment: intra-modality consistency is enforced by guiding each channel to learn coherent semantics within the infrared modality, while inter-modality alignment is achieved by minimizing the distribution gap between corresponding channels (e.g., R-to-R) across visible and infrared modalities. From the perspective of spectral modeling, Fan et al. \cite{fan2020cross} propose a cross-spectrum image generation method that synthesizes multiple spectral styles (e.g., blue, red, yellow, and grayscale) to encourage the model to learn shared features under diverse spectral conditions, thereby enhancing cross-modal generalization. Tan et al. \cite{tan2024rle} further explore the local linear response of materials to spectral variations and propose a dual-stage augmentation strategy that combines global channel perturbation with localized random linear enhancement to simulate realistic modality shifts. Alehdaghi et al. \cite{alehdaghi2025adaptive} point out that the lack of explicit objective modeling in intermediate image generation limits its adaptability. To address this, they propose an adversarial training framework that not only generates modality-bridging intermediate images but also explicitly guides the model to focus on learning identity-discriminative features.

\textbf{Feature Learning for Modality-Invariant Representation:}
  Motivated by the objective of isolating modality-related variations while preserving
  discriminative identity cues, feature disentanglement methods aim to explicitly separate
  modality-specific and modality-invariant factors, thus enabling more effective modality
  alignment.
From a broader perspective, these methods address a fundamental challenge in
  VI-ReID: visible and infrared images contain both shared identity information and modality-
  dependent factors, but the two types of information are difficult to separate perfectly.
  Existing studies therefore explore different forms of factorization, including identity-related
  and identity-irrelevant information, camera-specific and camera-agnostic features, and
  appearance-related and structural representations.

  Choi et al. \cite{choi2020hi} first introduce this idea into VI-ReID by disentangling identity-
  relevant factors (e.g., body shape, clothing) from identity-irrelevant ones (e.g., pose,
  illumination).
  Pu et al. \cite{pu2020dual} employ a Dual Gaussian-based Variational Auto-Encoder (DG-VAE) to
  divide features into identity-discriminative and identity-ambiguous subspaces.
  To mitigate intra-class discrepancies caused by style variations (e.g., background, exposure,
  color shifts) across cameras, Li et al. \cite{li2022camera} explicitly decompose features into
  two branches: a camera-specific branch that learns background features tied to camera
  attributes, and a camera-agnostic branch that removes style interference while retaining primary
  identity information.
Although these methods differ in their decomposition mechanisms, they share
  the same principle of preventing modality-specific variations from dominating the identity
  representation. Their main difference lies in the source of the nuisance factors being modeled:
  some methods focus on illumination and appearance, whereas others explicitly account for camera
  style or ambiguous identity information. This distinction is important because camera-related
  variations may remain within the same modality, while illumination and spectral differences are
  particularly prominent across modalities.
  Building on this concept, Zhang et al. \cite{zhang2022fmcnet} utilize disentangled identity
  features to generate missing-modality representations, training these in conjunction with
  original features. This approach, distinct from image-level generation, circumvents artifacts
  and structural distortions, thereby enhancing discriminability and stability in feature
  learning.
  This feature-level completion strategy also reveals an important evolution of
  disentanglement-based methods. Instead of using factorized features only to suppress irrelevant
  information, the learned identity-related representation is further used to compensate for
  missing modality information. In this way, feature disentanglement serves both as an alignment
  mechanism and as a basis for cross-modal feature reconstruction.

  To address the over-reliance on body shape in cross-modal representation learning, Feng et al.
  \cite{feng2023shape} introduce a shape-erased feature learning paradigm that eliminates body-
  shape information, compelling the model to capture a broader range of modality-shared features,
  which results in improved performance in visible-infrared person re-identification. Huang et al.
  \cite{huang2023exploring} expand upon the extraction of modality-shared appearance features by
  incorporating modality-invariant relational features among different body parts, thereby
  enhancing the robustness and generalization of cross-modal representations while maintaining
  discriminative appearance information. More recently, Zhou et al. \cite{zhou2024progressive}
  proposed a progressive discriminative feature learning strategy that implements varying
  alignment schemes at different stages to effectively extract discriminative features.
  These studies further suggest that modality-invariant information is not
  limited to global appearance or body shape. Relational structures among body parts and
  progressively learned semantic cues can also provide stable identity evidence. However,
  explicitly removing a potentially useful factor, such as body shape, may reduce identity
  discrimination for identities that are strongly characterized by body structure. Therefore,
  disentanglement methods need to balance the removal of modality-related bias with the
  preservation of identity-specific information.

  Beyond feature disentanglement, another line of research improves modality-
  invariant representation by optimizing the spatial and relational structure of feature
  learning.
  Early studies primarily aim at learning global shared representations across modalities
  \cite{dai2018cross,Wu_2017_ICCV,feng2019learning,ye2019bi}. Global features provide a holistic
  description of person images and capture modality-shared semantics. However, they are highly
  sensitive to background noise and non-target regions, which reduces the precision of identity
  representations. In challenging scenarios such as occlusion, complex backgrounds, or pose
  variations, global representations often fail to highlight the most discriminative regions,
  thereby limiting the robustness and discriminability of cross-modal matching.
  The limitation of global representation learning is therefore not simply
  insufficient feature capacity, but the difficulty of distinguishing identity-related regions
  from background and modality-dependent regions. This motivates local feature learning as a
  complementary strategy.
 To address these limitations, local feature learning is introduced as a complementary strategy,
  since it focuses on fine-grained and region-specific cues that are less affected by background
  interference. Within this direction, Ye et al. \cite{ye2020dynamic} propose a Dynamic Tri-level
  Relation Mining (DTRM) framework that models cross-modal semantic relations at three levels,
  namely channel, part, and graph structure, in order to achieve joint optimization. Zhu et al.
  \cite{zhu2020hetero} introduce the Two-Stream Local Feature Network (TSLFN) to emphasize fine-
  grained local feature learning. Zhang et al. \cite{zhang2021global} further integrate global and
  multi-scale local features, enabling complementary and mutually reinforced representations
  across modalities, which enhance both robustness and discriminability in cross-modal matching.
  The progression from global modeling to local and relational modeling
  reflects an increasing effort to capture identity cues at multiple spatial scales. DTRM
  emphasizes structured relations among feature elements, TSLFN focuses on fine-grained body
  regions, and the global-local fusion strategy combines holistic identity semantics with detailed
  local evidence. Nevertheless, local alignment may be affected by pose variation, occlusion, and
  inaccurate body-part correspondence, which limits the reliability of purely local matching.

  Park et al. \cite{park2021learning} establish pixel-level correspondences between RGB and IR
  images through a module called CMAlign, which avoids noisy matches caused by conventional coarse
  image-level alignment and guides the model to learn more discriminative identity features. Yang
  et al. \cite{yang2022learning} address the simultaneous presence of noisy identity annotations
  and incorrect cross-modal correspondences, collectively referred to as Twin Noisy Labels (TNL),
  by integrating sample cleanliness estimation and a pairwise correction mechanism. This approach
  effectively mitigates the negative impact of both annotation noise and matching noise.
  These methods show that improving feature quality also requires improving the
  reliability of cross-modal correspondence. Pixel-level alignment attempts to obtain more precise
  spatial matching, whereas TNL explicitly considers the fact that both identity labels and
  modality correspondences may be unreliable. Thus, the effectiveness of local and relational
  feature learning depends not only on the representation architecture but also on the quality of
  the supervision used to establish cross-modal relationships.

  Additional research, such as Zhang et al. \cite{zhang2022modality}, proposes to fuse the fine-
  grained advantages of visible images with the stable characteristics of infrared images for
  joint representation learning. Based on the fused features, the model further performs refined
  enhancement by exploiting the complementary strengths of both modalities. These studies
  \cite{hao2021cross,yu2023modality,wei2021syncretic} realize modality confusion at the feature
  level by incorporating rich semantic information from both visible and infrared images.
  Unlike image-level generation, feature-level modality confusion directly
  constrains the representation used for retrieval. Its underlying assumption is that modality
  identity should be weakened in the shared feature space while identity information should remain
  discriminative. This reduces the risk of generation artifacts and avoids the additional
  computational cost of synthesizing images. However, overly strong modality confusion may also
  remove useful modality-specific cues, resulting in a trade-off between modality invariance and
  identity discrimination.

  Compared with traditional GAN-based image generation methods, such feature-level confusion
  mechanisms avoid generation noise and additional computational overhead, thereby enabling more
  efficient learning of modality-invariant representations. Therefore, feature-level fusion and modality confusion can be viewed as
  complementary to feature disentanglement: disentanglement explicitly separates different
  factors, whereas modality confusion directly regularizes the shared representation to reduce
  modality predictability.

  With the rise of Transformer architectures, their potential in VI-ReID has been increasingly
  investigated. Liang et al. \cite{liang2021cmtr} are the first to introduce Transformer networks
  into the cross-modal Re-ID task, leveraging their strong capability for global modeling to
  extract richer and more discriminative features. Subsequently, Zhao et al.
  \cite{zhao2022spatial} demonstrate that Transformers are not only effective in modeling global
  interactions among image patches, but also exhibit strong potential in capturing sequence
  correlations at both the pixel and channel levels. Accordingly, they propose a method that
  models long-range dependencies across spatial and channel dimensions, thereby enhancing the
  discriminative power of features in cross-modal person re-identification. The dual-stream
  Transformer architecture proposed in \cite{chai2023dual} addresses the limitation of CNNs in
  modeling global dependencies, enabling the capture of long-range dependencies in pedestrian
  images.
  These studies indicate that Transformer-based architectures provide a
  flexible way to integrate global context, local patch interactions, and cross-modal dependencies
  within a unified representation. However, the self-attention mechanism alone does not guarantee
  modality invariance. Without an appropriate modality-sharing or modality-confusion strategy,
  long-range modeling may also propagate modality-specific noise across the feature space. Thus,
  the key issue is not only whether Transformers can model global dependencies, but also how
  shared and modality-specific information should be controlled during this process.

  Most methods for identity representation primarily focus on visual information, relying heavily
  on appearance features. However, this reliance renders models vulnerable to occlusion, pose
  variations, and background clutter, which impedes the extraction of identity-discriminative
  features. To mitigate these challenges, an increasing number of studies incorporate auxiliary
  information such as pose estimation, body keypoints, attribute annotations, or semantic
  segmentation to provide structural guidance for feature learning.
  The motivation of this direction is to introduce relatively stable structural
  or semantic cues that are less sensitive to illumination and spectral changes. Such auxiliary
  information does not replace visual appearance, but provides additional constraints for
  identifying body structure, local regions, and semantic attributes across modalities. For example, the Pose-guided Modality-invariant Feature Alignment method \cite{liu2024pose} uses
  human posture as a structural clue to uniformly align the features of visible light and infrared
  images; in addition, auxiliary posture estimation tasks \cite{miao2023exploring} further verify
  the effectiveness of key point position information as a modality sharing clue in cross-modal
  ReID. On the other hand, Wei et al. \cite{wei2021flexible} propose a method called FBP-AL
  (Flexible Body Partition with Adversarial Learning), which automatically partitions human body
  parts and integrates structure-assisted supervision with weighted feature fusion to guide the
  model in capturing identity-related local features more effectively.
  These pose- and body-structure-based methods are particularly useful when
  color and texture information becomes unreliable. Nevertheless, their performance may depend on
  the accuracy of pose estimation or body-part partitioning, which can be degraded by occlusion,
  low resolution, and imperfect infrared imaging. This illustrates the general trade-off of
  auxiliary information: stronger structural guidance can improve alignment, but inaccurate
  auxiliary cues may introduce additional noise.

  In addition to pose information, researchers also explore other forms of auxiliary information.
  Wang et al. \cite{wang2022cris} employ the CLIP model with learnable bimodal text tags to encode
  modality-specific semantics in visible and infrared images and fuse them into a unified
  representation, thereby enhancing the modeling of modality complementarity. Yu et al.
  \cite{yu2025clip} propose a Semantic Margin-guided Feature Decoupling (SMFD) module that
  decomposes image features into identity-related and style-related components, with text
  embedding constraints imposed to strengthen the semantic consistency of the former and achieve
  more accurate cross-modal alignment. Lee et al. \cite{lee2023camera} develop a camera-aware
  progressive adaptation framework that leverages camera-specific labels of pedestrian images to
  incrementally transfer knowledge from the source domain to the target domain, thereby improving
  generalization across cameras and modalities. Jiang et al.~\cite{jiang2024joint} innovatively
  introduce natural language guidelines to learn both global and local contextual information for
  VI-ReID, uncovering additional latent cues.
  These methods extend auxiliary guidance from physical structure to semantic
  and contextual knowledge. Text-based methods provide high-level descriptions that can supplement
  incomplete visual evidence, whereas camera-aware methods explicitly model the domain factors
  responsible for feature shifts. Their common objective is to constrain the learned
  representation with information that is more stable or interpretable than raw appearance. At the
  same time, the dependence on pretrained models, semantic annotations, or camera labels may limit
  their applicability in settings where such information is unavailable.

  In summary, the above studies address modality-invariant representation
  learning from complementary perspectives. Feature disentanglement separates identity-related
  factors from modality- or camera-specific variations; global-local and relational modeling
  improves the spatial precision of identity representations; feature-level modality confusion
  directly reduces the modality discrepancy in the embedding space; Transformer architectures
  enhance long-range dependency modeling; and auxiliary information provides additional structural
  or semantic constraints. These strategies are not mutually exclusive, but their combination
  requires a careful balance between modality invariance, identity discrimination, correspondence
  reliability, computational cost, and dependence on external priors.

\textbf{Metric Learning for cross-modal Alignment:}
Alongside feature learning, metric learning also plays a vital role in VI-ReID by enforcing discriminative constraints in the embedding space, encouraging intra-class compactness and inter-class separability, and thereby enhancing identity discrimination across modalities.
Traditional metric learning in ReID primarily relies on distance-based constraints between samples. Identity loss ensures correct classification of each instance, contrastive loss encourages cross-modal alignment by pulling intra-class samples closer and pushing inter-class samples farther apart, and triplet loss constructs anchor–positive–negative triplets to guarantee that positive samples of the same identity are closer to the anchor than negative samples of different identities. 
To further improve discriminability, advanced extensions such as quadruplet loss \cite{chen2017beyond} , center loss \cite{wen2016discriminative} and hetero-center triplet loss \cite{liu2020parameter}, and cluster contrast loss \cite{cheng2023cross} are also introduced.
In addition to these classical losses, recent studies design novel loss functions tailored for cross-modal scenarios. For instance, Jia et al. \cite{jia2020similarity} observe that existing works often overemphasize modality discrepancies while ignoring intra-modality similarities, and thus propose the Similarity Inference Metric (SIM) to leverage intra-modality similarity for alleviating the modality gap. Zhang et al. \cite{zhang2022hybrid} introduce a hybrid constraint framework that jointly optimizes intra- and inter-modality feature distributions by enforcing both class-level and modality-level similarity constraints. Gao et al. \cite{mso} further present a novel approach that integrates Perceptual Edge Feature (PEF) loss with cross-modal Contrastive Center (CMCC) loss, enabling joint optimization to enhance feature discriminability.
Kong et al. \cite{kong2021dynamic} propose Dynamic Center Aggregation Loss, which fuses mixed-modality information and dynamically adjusts center distances between different modalities, thereby achieving effective constraints on multimodal distribution relationships. Cai et al. \cite{cai2021dual} strengthen category aggregation and separation by learning category centers in visible and infrared modalities separately, and construct triplets based on a dynamic hard sample mining strategy.
Subsequently, Cheng et al. \cite{cheng2023cross} design a deep comparative learning framework with an aggregated memory mechanism to enhance inter-modal alignment by modeling cross-modal semantic consistency. Most recently, Kim et al. \cite{kim2024enhancing} introduce Positive/Negative Enhancement Loss, which expands the representation space of the same identity within a single modality to improve feature diversity and discrimination, thereby further promoting generalization across modalities.

\subsubsection{Unsupervised Visible-Infrared Person ReID}

\begin{figure}
    \centering
    \includegraphics[width=1\linewidth]{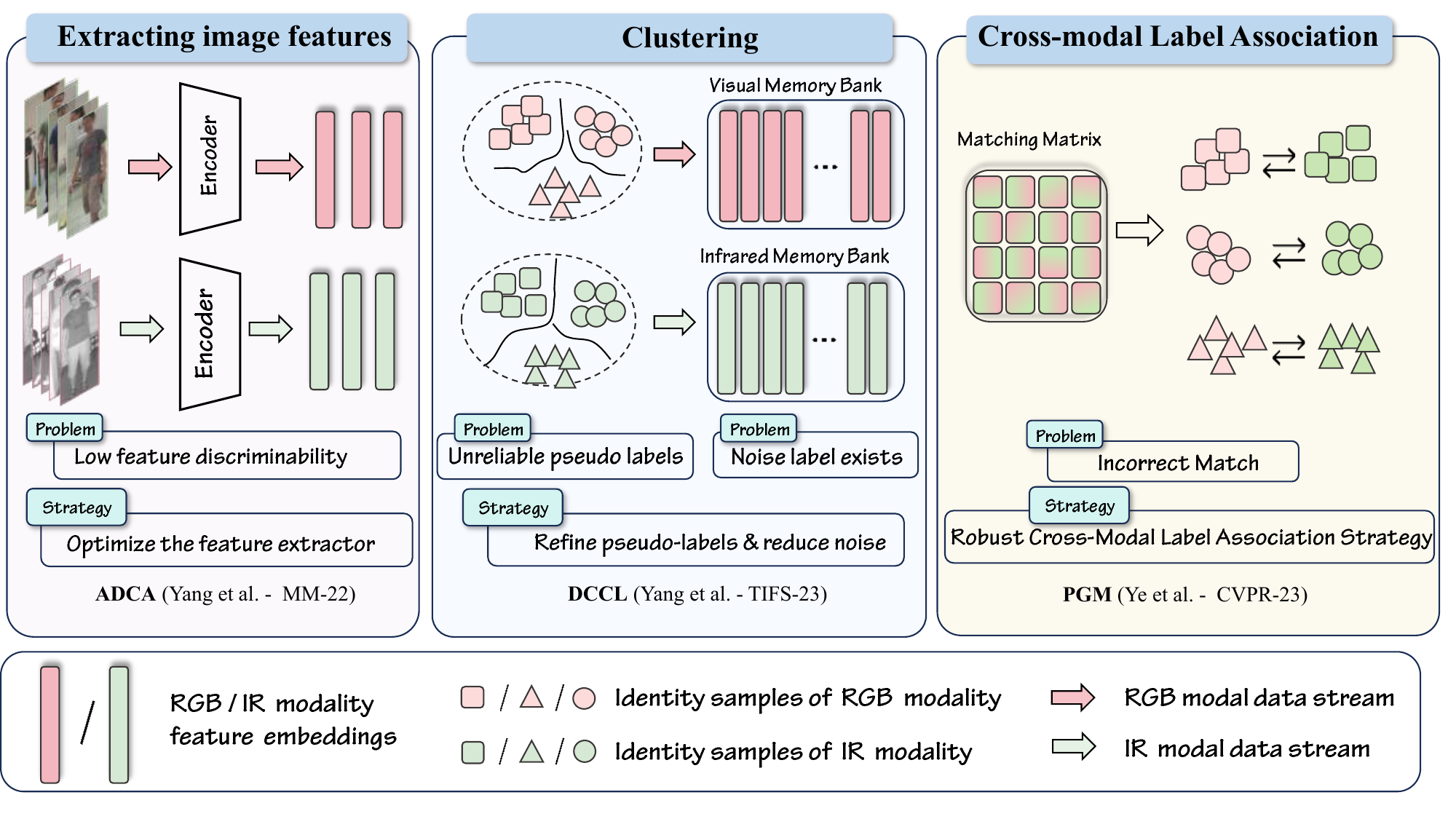}
    \caption{Overview of unsupervised visible-infrared person re-identification, covering three main research aspects: feature extraction, clustering, and label association. The figure also highlights the major challenges and representative strategies within each aspect.}
    \label{fig:unsup}
\end{figure}

Recent advancements in supervised visible-infrared person re-identification (VI-ReID) have achieved high recognition accuracy, demonstrating the effectiveness and maturity of cross-modal modeling. However, practical scenarios often involve infrared images captured under nighttime or low-light conditions, leading to low clarity and making manual annotation challenging. Additionally, the inherent costs of cross-modal pairing limit scalability across heterogeneous devices and diverse real-world environments, thereby constraining the generalizability and deployment feasibility of supervised methods.
To mitigate dependence on labeled data and enhance real-world applicability, there is increasing focus on unsupervised visible-infrared person re-identification (US-VI-ReID). This approach aims to learn pedestrian representations that bridge the gap between visible and infrared modalities without the need for manually annotated identity labels, facilitating cross-modal image retrieval. By circumventing the costly and time-consuming process of cross-modal annotation, US-VI-ReID methods are better suited for deployment in large-scale, open scenarios and represent a crucial direction for advancing cross-modal ReID applications. We illustrate the pipeline for unsupervised VI-ReID in Fig. \ref{fig:unsup} and provide a comparative analysis of existing methods in Table \ref{tab:unsup}.



Based on these challenges, existing studies mainly explore three key dimensions: (1) cross-modal invariant feature learning, which aims to directly model modality-shared identity features and reduce inter-modal discrepancies; (2) cross-modal pseudo-label generation and enhancement, which focus on producing more reliable pseudo-labels to guide feature learning while mitigating the adverse effects of noisy labels; and (3) cross-modal label association, which establishes identity correspondences between visible and infrared samples to provide cross-modal supervisory signals and enable the learning of robust identity representations.

\textbf{Cross-modal Invariant Feature Learning:}
Modality-invariant feature learning centers on the core objective of extracting identity-discriminative yet modality-agnostic representations from heterogeneous inputs. The key challenge lies in reducing the distributional discrepancy between visible and infrared images in the feature space, while preserving the structural and semantic information necessary for person identification. Liang et al. ~\cite{liang2021H2H} pioneers the first unsupervised framework H2H, which employs a two-phase strategy. A homogeneous learning phase generating intra-modality pseudo-labels and a heterogeneous learning phase aligning cross-modal features to extract invariant representations. Yang et al. \cite{yang2022ADCA} advances this study by proposing the ADCA framework, which enhances feature invariance through augmented dual-contrastive learning and cross-modal memory aggregation.
To tackle hierarchical discrepancies (intra-camera, inter-camera, and cross-modal variations) in cross-modal retrieval, the GUR framework \cite{yang2023GUR} achieves discrepancy disentanglement and robust matching via hierarchical representation learning. 
\textbf{More recently, SDCL~\cite{yang2024SDCL} extends modality-invariant feature learning by jointly exploiting shallow and deep feature representations. Instead of relying solely on the final-layer representation, its shallow--deep collaborative learning strategy leverages complementary structural details and high-level semantic cues, leading to more robust modality-invariant representations for unsupervised VI-ReID.}

\textbf{Cross-modal pseudo-label generation and enhancement:}
In unsupervised visible-infrared person re-identification, pseudo labels function as surrogate supervisory signals, critically influencing the feature space structure, cross-modal alignment, and the stability of label propagation and matching. Consequently, substantial research has concentrated on the reliable generation and refinement of pseudo labels.
Given that clustering underpins pseudo label generation, several approaches aim to enhance reliability by optimizing the clustering process, promoting feature discriminability and structural consistency to mitigate error accumulation during training. For instance, Lin et al. \cite{lin2019bottom} propose a bottom-up clustering (BUC) method that initially treats each sample as an independent identity to maximize inter-identity diversity, subsequently merging similar samples to enhance intra-identity similarity. Similarly, CHCR ~\cite{pang2023CHCR} introduces a hierarchical clustering strategy that constructs stable local structures within each modality and performs cluster-level alignment across modalities, progressively generating more reliable cross-modal positive pairs.
Despite these advancements, existing methodologies often overlook the impact of intra-modality camera differences. Variations in appearance due to multi-camera setups (cam-style variation) can exacerbate feature shifts within a modality, resulting in inconsistent clustering. To address this challenge, several studies have integrated camera labels to facilitate more accurate clustering \cite{li2022camera,li2024inter,yang2025dynamic,xia2025camera,wu2025extended,yang2023GUR}.


In traditional methods (e.g., \cite{liang2021H2H,yang2022ADCA,wu2023PGM,wang2022OTLA}), pseudo-label assignment typically relies on offline clustering, i.e., clustering is performed periodically on extracted features rather than being updated continuously during training. However, such static label associations are prone to error accumulation during training, which can hinder model optimization. To alleviate this issue, Yang et al. \cite{yang2023DCCL} proposes a collaborative pseudo-label refinement framework that integrates both offline and online mechanisms. Specifically, an online cross-modal feature memory module is introduced, comprising Instance Memory, Identity Memory, and Domain Memory. This dynamic update strategy mitigates the impact of noisy labels by smoothing label transitions and has inspired subsequent research.
Shi et al. \cite{shi2025MMM} further extends the memory mechanism by decomposing the conventional unified memory into multiple sub-memories, enabling multi-view feature representation.
Traditional cross-modal clustering-based matching strategies adopt a one-to-one scheme, where each visible cluster is matched to a single infrared cluster. Given the limited information in infrared images, this often results in information asymmetry and insufficient matching. To address this issue, Cheng et al. \cite{cheng2023MBCCM} introduces an innovative Many-to-Many Bilateral Matching mechanism, which alleviates mismatches caused by inaccurate cluster boundaries.

Following the generation of initial pseudo-labels, several studies have focused on evaluating their posterior reliability and refining their utilization \cite{liu2024PRAISE, yin2025adaptive, dai2025dual}. These methods typically assess pseudo-label confidence after clustering to distinguish clean samples from noisy ones, and adopt quality-aware training strategies accordingly, thereby enhancing the robustness of pseudo-supervision and improving overall training stability.
Treating all hard samples—those distant from cluster centers—as noisy may lead to the misjudgment and discard of valuable supervisory signals. MIMR \cite{pang2024MIMR} argues that some hard samples are actually ``ambiguous samples'' located near the decision boundaries of multiple clusters, which can confuse model training. Therefore, it proposes to further categorize hard samples into ``normal hard samples'' and ``ambiguous samples'', and explicitly remove the latter to enhance the accuracy of pseudo-label learning.
Shi et al. \cite{shi2024learning} categorizes sample prototypes into three types—Centroid Prototype, Hard Prototype, and Dynamic Prototype—each serving different optimization objectives. A progressive contrastive learning strategy is employed: Centroid Prototypes are utilized in the early training stages to capture commonality, while Hard and Dynamic Prototypes are gradually incorporated later to model discriminative and diverse features.


In evaluating the importance and reliability of pseudo-labels, neighbor labels are often treated as a structural prior that provides auxiliary cues for local semantic consistency. Specifically, many studies \cite{cheng2023DOTLA, yin2024robust} leverage the neighborhood consistency assumption, which posits that if a sample’s predicted label is consistent with those of its neighbors in the feature space, it is more likely to be correct and reliable. Based on this principle, an increasing number of methods have introduced neighborhood structure modeling to assess pseudo-label confidence via neighbor label distributions and refine training strategies accordingly, thereby improving the stability and robustness of pseudo-supervised learning. In addition, Pang et al. \cite{pang2023CHCR} propose an inter-channel pseudo-label refinement strategy based on the assumption that all three channels of the same visible image should share a consistent pseudo-label. This approach filters noisy labels and enhances pseudo-label reliability. N-ULC \cite{teng2025relieving} replaces hard pseudo labels in homogeneous and heterogeneous spaces with neighbor-derived soft labels and dynamically downweights unreliable samples, thereby treating label confidence as a continuous quantity.

\textbf{Cross-modal Label Association:}
In US-VI-ReID, the development of cross-modal pseudo-labels is vital for modality alignment and feature learning. The PGM method \cite{wu2023PGM} utilizes modality-specific clustering graphs and cost matrices for cross-modal graph matching, but its cluster-level labeling can lead to incorrect groupings and neglect fine-grained instance-level semantics. Ye et al. \cite{ye2025dual} integrate instance-level matching and outlier filtering. Conversely, OTLA \cite{wang2022OTLA} reformulates label assignment as an optimal transport problem, employing Sinkhorn optimization for semantic alignment and balanced label distribution, thus mitigating visible modality dominance. Cheng et al. \cite{cheng2023DOTLA} build upon this with a bidirectional transport mechanism that facilitates mutual projections of visible and infrared images into each other's pseudo label space. Addressing the limitations of single memory modules for identity association, Shi et al. \cite{shi2025MMM} propose a multi-memory framework to capture subtle identity differences. Additionally, Yang et al. \cite{yang2025progressive} introduce a progressive learning paradigm that combines intra-modal adversarial learning, cross-modal neighbor expansion clustering, and modality-invariant contrastive optimization to refine pseudo labels and reduce noise propagation. Finally, Teng et al. \cite{teng2024enhancing} apply a bidirectional consistency rule to stabilize matching between modalities, minimizing early training noise accumulation. MCL \cite{yao2025unsupervised} studies an unpaired setting in which visible and infrared identity populations only partially overlap, rather than assuming full cross-modal identity correspondence. It uses Cross-modality Feature Mapping to construct a
pseudo-identity space and Static-Dynamic Collaborative learning to align cluster and instance-level correspondences under identity mismatch, this setting reduces implicit
pairing supervision.

\begin{table*}[t]
\centering
\setlength{\tabcolsep}{2pt} 
\footnotesize
\setlength{\tabcolsep}{7pt}
\caption{Performance comparison of unsupervised VI-ReID methods on SYSU and RegDB datasets in terms of Rank-1 accuracy and mAP.}
\begin{tabular}{l|l|cc|cc|cc|cc}
\toprule
\multirow{3}{*}{} & \multirow{3}{*}{} & \multicolumn{4}{c|}{SYSU}  & \multicolumn{4}{c}{RegDB}  \\
\cmidrule{3-10}
\multirow{1}{*}{Methods} & \multirow{1}{*}{Venue} & \multicolumn{2}{c|}{All Search} & \multicolumn{2}{c|}{Indoor Search} & \multicolumn{2}{c|}{VIS to IR} & \multicolumn{2}{c}{IR to VIS}  \\
 & & R-1  & mAP & R-1  & mAP & R-1  & mAP & R-1  & mAP  \\
\hline
H2H \cite{liang2021H2H}&TIP-21& 30.15  & 29.40 & - &  - & 23.81 & 18.87 & -  & - \\
OTLA \cite{wang2022OTLA}&ECCV-22 & 29.9 & 27.1 & 29.8  & 38.8 & 32.9  & 29.7 & 32.1  & 28.6 \\
ADCA \cite{yang2022ADCA}&MM-22 & 45.51  & 41.73 & 50.60 & 59.11  & 67.20 & 64.50 & 68.48 & 63.81 \\
CHCR \cite{pang2023CHCR}&TCSVT-23 & 59.47 & 59.14 & -  & - & 69.31 & 64.74 & 69.96 & 65.87 \\
DOTLA \cite{cheng2023DOTLA}&MM-23 & 50.36  & 47.36 & 53.47 & 61.73 & 85.63 & 76.71 & 82.91 & 74.97 \\
MBCCM \cite{cheng2023MBCCM}&MM-23 & 53.14 & 48.16 & 55.21 & 61.98 & 83.79 & 77.87 & 82.82 & 76.74 \\
PGM \cite{wu2023PGM}&CVPR-23 & 57.27 & 51.78 & 56.23 & 62.74 & 69.48 & 65.41  & 69.85 & 65.17  \\
DCCL \cite{yang2023DCCL}&TIFS-23 &63.81 & 58.62 & 66.67 & 71.82 & 78.28 & 71.98 & 78.28 & 71.30  \\
GUR \cite{yang2023GUR}&ICCV-23 & 63.51 & 61.63 & 71.11 & 76.23 & 76.91 & 70.23 & 75.00 & 69.94  \\
MIMR \cite{pang2024MIMR}&KBS-24 & 58.69 & 58.16 & 65.39 & - & 66.93 & - & 66.43 & - \\
CMCP \cite{teng2024enhancing}&MM-24 &61.7 & 56.1 & 60.9 & 66.5 & 86.8 & 81.7 & 86.7 & 82.3 \\
CMAM \cite{wu2024CMAM}&TCSVT-24&62.0 & 58.2 & 67.6 & 72.7 & 89.1 & 74.0 & 89.0 & 74.0 \\
PCLHD \cite{shi2024learning}&NeurIPS-24 &64.4 & 58.7 & 69.5 & 74.4 & 84.3 & 80.7 & 82.7 & 78.4 \\
SDCL \cite{yang2024SDCL}&CVPR-24& 64.49 & 63.24 & 71.37 & 76.90 & 86.91 & 78.92 & 85.76 & 77.25 \\
PCAL \cite{yang2025progressive}& TIFS-25& 54.39 & 51.95 & 59.69 & 66.72 & 86.43 & 82.51 & 86.21 & 81.23 \\
N-ULC \cite{teng2025relieving}& AAAI-25& 61.81 & 58.92 & 67.04 & 73.08 & 88.75 & 82.14 & 88.17 & 81.11 \\
MCL \cite{yao2025unsupervised} &ICCV-25& 62.95 & 62.71 & 67.81 & 74.19 & 89.83 & 83.12 & 88.64 & 82.04 \\

\bottomrule
\end{tabular}
\label{tab:unsup}
\end{table*}

\subsection{Text-Image ReID}



\begin{figure}
    \centering
    \includegraphics[width=1\linewidth]{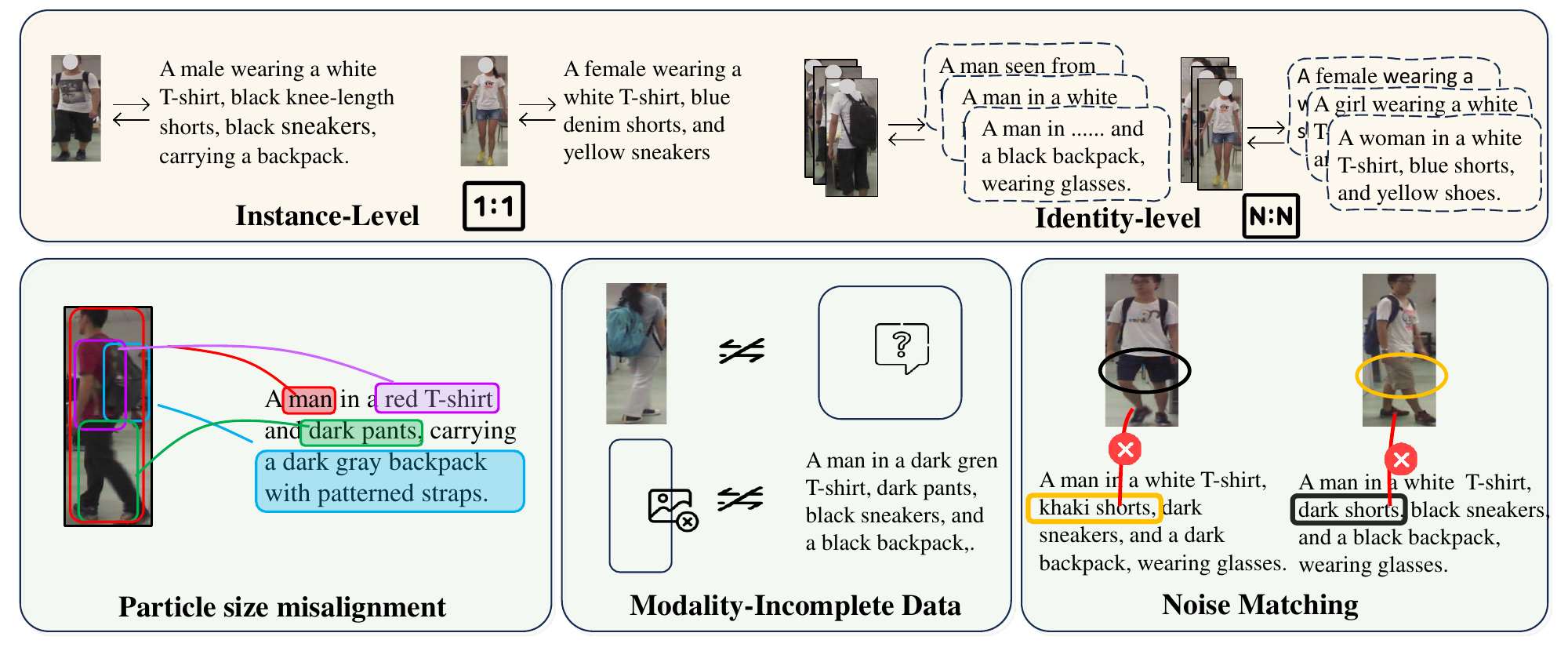}
    \caption{This figure contrasts instance-level (1:1) and identity-level (N:N) text-to-image matching in TI-ReID (top), and outlines primary research challenges such as particle size misalignment, modality-incomplete data, and noise matching (bottom).}
    \label{fig:TI-ReID}
\end{figure}

In real-world surveillance and public security scenarios, complete visual information of the target individual is often unavailable as a query reference. In contrast, verbal descriptions provided by eyewitnesses—such as clothing attributes, body shape, and carried items—are more common and accessible. Therefore, how to accurately retrieve the corresponding pedestrian image from a large-scale gallery based solely on such textual information has become a research problem of both practical importance and urgency. This has led to the emergence of the Text-Image Person Re-Identification (TI-ReID).
TI-ReID aims to retrieve target pedestrians by using natural language descriptions as queries under cross-modal conditions. Compared with conventional image-based ReID, TI-ReID encounters more severe cross-modal discrepancies. The text modality is typically abstract, flexible, and highly subjective in nature, while the image modality provides concrete, continuous, and structurally rich visual information. The inherent differences in representation form and semantic level between the two modalities give rise to a significant modality gap, posing considerable challenges for achieving effective cross-modal feature alignment and consistent identity matching. We illustrate the primary challenges encountered in TI-ReID in Fig. \ref{fig:TI-ReID} and compare the performance of existing methods in Table \ref{tab:ti-reid}.

\subsubsection{Datasets}

CUHK-PEDES \cite{li2017person} is a text-based person retrieval dataset proposed by the Chinese University of Hong Kong and is one of the most widely used benchmarks in the field of Text-Image ReID. It is constructed by integrating image resources from CUHK03 \cite{CUHK03}, Market-1501 \cite{Market-1501}, SSM \cite{SSM}, VIPeR \cite{VIPeR}, and CUHK01 \cite{CUHK01}, supplemented with natural language descriptions collected via the Amazon Mechanical Turk (AMT) platform. The dataset contains 40,206 images of 13,003 identities paired with 80,412 textual descriptions (each with no fewer than 23 words), and is divided into a training set (34,054 images, 11,003 identities) and validation/test sets (about 3,000 images and 6,100 descriptions associated with 1,000 identities, respectively). This dataset is characterized by rich semantic diversity, with descriptions covering various aspects such as appearance, posture, actions, and interactions with objects.

ICFG-PEDES \cite{ding2107semantically} is a person description dataset constructed by South China University of Technology, designed to facilitate research in fine-grained person description and image-text matching. The dataset comprises 54,522 pedestrian images corresponding to 4,102 distinct identities, all collected from the MSMT17 \cite{MSMT17} dataset. Each image is annotated with a single natural language description, with an average length of 37.2 words. The training set contains 34,674 image-text pairs associated with 3,102 identities, while the test set consists of 19,848 pairs covering 1,000 identities.
Compared to CUHK-PEDES, ICFG-PEDES places greater emphasis on fine-grained descriptions of pedestrian appearance, and deliberately removes irrelevant background and redundant information to improve semantic clarity in image-text alignment. Furthermore, the dataset construction introduces greater appearance diversity to mitigate the background consistency bias observed in previous datasets. As a result, ICFG-PEDES serves as a more challenging benchmark for evaluating model performance under complex and detail-sensitive matching scenarios. Table \ref{tab:datasets-TIReID} summarizes the detailed statistics of the datasets.

RSTPReid \cite{zhu2021dssl} is a text-image person re-identification dataset constructed by Nanjing University of Science and Technology, aiming to address the limitations of CUHK-PEDES, where each pedestrian is typically captured by the same camera under similar time-space conditions—an unrealistic setting for real-world applications. Built upon the MSMT17 \cite{MSMT17} dataset, RSTPReid contains 20,505 images of 4,101 identities, captured by 15 distinct cameras under diverse viewpoints, lighting conditions, locations, and weather scenarios, offering significantly higher environmental variability and realism. Each identity in the dataset is represented by five images captured from different camera perspectives, accompanied by two natural language descriptions, each containing no fewer than 23 words. The dataset is partitioned into a training set comprising 3,701 identities, while the validation and test sets each consist of 200 identities. Despite the relatively limited number of identities, RSTPReid encompasses both indoor and outdoor environments, highlighting significant variations in camera angles and illumination conditions. This characteristic renders it a more rigorous and representative benchmark for Text-Image ReID tasks. Comprehensive details regarding the datasets are presented in Table \ref{tab:datasets-TIReID}.


\begin{table*}[htbp]
  \caption{Statistics of TI-ReID related datasets, including the number of identities, RGB images, and textual descriptions.}
  \label{tab:datasets-TIReID}
  
  \centering
  {\fontsize{8pt}{10pt}\selectfont 
  \begin{tabular*}{\textwidth}{@{\extracolsep{\fill}}cccc@{}}
    \toprule
    Datasets & Identities & RGB Imgs & Texts \\
    \midrule
    CUHK-PEDES \cite{li2017person} & 13,003 & 40,206 & 80,412 \\
    ICFG-PEDES \cite{ding2107semantically}  & 4,102 & 54,522  & 54,522 \\
    RSTPReid \cite{zhu2021dssl}  & 4,101 & 20,505 & 41,010  \\
    \bottomrule
  \end{tabular*}}
\end{table*}

\textbf{Summary:}Overall, these benchmarks differ substantially in data provenance and annotation strategy, leading to distinct benchmark biases. CUHK-PEDES is constructed by integrating multiple ReID datasets, resulting in rich semantic descriptions but also introducing background and scene consistency biases. In contrast, ICFG-PEDES and RSTPReid, both built upon MSMT17, place greater emphasis on fine-grained appearance descriptions and cross-camera diversity, respectively, providing more realistic and challenging evaluation settings. These differences should be taken into account when comparing model performance across benchmarks.

\subsubsection{Methods}


At the early stage of research on Text-Image ReID (TI-ReID), most approaches employed a global matching paradigm \cite{li2017identity,zhang2018deep,chen2018improving,ye2018hierarchical,chen2021cross,wu2021lapscore,wu2023refined}, where entire images are mapped to comprehensive textual descriptions within a unified semantic space, thereby enhancing the efficacy of cross-modal retrieval.

However, natural language descriptions are typically concise and semantically dense, containing numerous fine-grained cues that are crucial for identity discrimination (e.g., “carrying a black shoulder bag”, “wearing a blue hat”). When relying solely on global representations, models tend to overemphasize semantically strong but low-discriminative words (e.g., “man”, “person”), while neglecting key descriptions that truly convey identity differences. Moreover, in real-world scenarios such as surveillance, eyewitness recollections of suspects often focus on salient attributes and local details, further highlighting the importance of fine-grained alignment modeling \cite{chen2022tipcb}\cite{gao2021contextual}. 
Moreover, \cite{chen2018improving}\cite{qi2023image} jointly leverage global identity-level descriptions and local phrase-level region features to construct more fine-grained text-image alignment. Shao et al. \cite{shao2023unified} propose a unified pre-training framework specifically designed for the TI-ReID task. To support this framework, they also construct a large-scale text-annotated dataset, LUPerson-T, which achieves consistent alignment between image and text modalities at both data and training levels.
Han et al. \cite{han2021text} propose a novel cross-modal momentum contrastive learning framework to better learn more discriminative feature representations on small-scale datasets.

 In addition, images and text differ significantly in structural representation and semantic granularity. For instance, the word “man” in text may correspond to the entire human figure in an image, while a local image region (e.g., a backpack) may align with a more detailed phrase such as “carrying a blue backpack.” Moreover, images typically contain rich visual information, whereas textual descriptions are often coarse, leading to granularity mismatches that hinder precise cross-modal alignment and may result in semantic drift or retrieval failures.
To address this issue, some studies introduce explicit granularity alignment mechanisms. For example, Jing et al. \cite{jing2020pose} align full-sentence descriptions with relevant image regions and leverage human pose information to guide the alignment between body parts in images and corresponding noun phrases in text, thereby bridging semantic granularity gaps. Niu et al. \cite{niu2020improving} propose a multi-granularity alignment framework that sequentially aligns image and text at three levels: global-global (entire image with entire text), global-local (entire image with text phrases), and local-local (image regions with text segments), progressively refining the matching process. Shao et al. \cite{shao2022learning} propose the LGUR framework, which employs a dictionary-guided alignment module (DGA) and a prototype-guided unification module (PGU) to explicitly align visual regions with textual semantic units in the feature space, effectively mitigating granularity discrepancies and enhancing both matching accuracy and cross-domain generalization. Additionally, Huang et al. \cite{huang2024cross} extract multi-scale semantic features from both text and image using variable-sized windows and project them into a shared space to capture alignment across different semantic levels.

Most existing studies leverage uni-modal pre-trained models as backbone encoders; for example, Long Short-Term Memory (LSTM) networks or BERT \cite{devlin2019bert} are employed for textual encoding, while ResNet \cite{he2016deep} or Vision Transformer (ViT) \cite{dosovitskiy2020image} are utilized for visual encoding. However, these approaches often fail to fully exploit the synergistic potential inherent in the interaction between different modalities.
In recent years, Visual-Language Pre-training (VLP) methods have garnered increasing attention due to their superior capability in learning cross-modal representations. Among them, CLIP\cite{radford2021learning} stands out as a representative framework, constructing a unified multimodal semantic space via large-scale contrastive learning on image-text pairs. A large number of studies have fine-tuned CLIP for various downstream tasks, such as text-video retrieval \cite{luo2021clip4clip,fang2021clip2video,zhao2022centerclip}, referring image segmentation\cite{wang2022cris}, and general video recognition \cite{ni2022expanding}, achieving remarkable performance across these applications. Inspired by this trend, researchers have begun to explore the potential of CLIP in the TI-ReID task, aiming to leverage its strong cross-modal alignment capabilities to enhance text-based person retrieval performance.


Li et al. \cite{li2023clip} represents the first systematic attempt to introduce CLIP into ReID tasks, in which the authors develop a two-stage training framework named CLIP-ReID.
Yan et al. \cite{yan2023clip} propose CFine, a fine-grained information mining framework based on CLIP, which aims to transfer the knowledge of CLIP to the TI-ReID task and achieve fine-grained text-image alignment.
It is noteworthy that CLIP’s pretraining objective and data domain are inherently misaligned with the requirements of the TI-ReID task. 
 Cao et al. \cite{cao2024empirical} conduct a comprehensive study on CLIP’s performance in TI-ReID from the perspectives of data augmentation and loss design. By incorporating tailored training heuristics, augmentation strategies, and customized loss functions, they establish a strong TBPS-CLIP baseline tailored to the TIReID setting.
CLIP is trained on large-scale image-text datasets; however, its training domain differs significantly from that of TI-ReID, with image scenes and textual description styles being entirely distinct. Directly applying CLIP to TI-ReID may restrict the model’s discriminative ability and generalization performance in identity-level retrieval. To address this issue, Li et al. \cite{li2024prompt} propose a two-stage training strategy. In the first stage, only prompts are optimized with contrastive loss to achieve domain adaptation, enabling the model to better align with the TI-ReID data distribution. In the second stage, with prompts fixed, the encoders are fine-tuned to focus on fine-grained identity features, thereby accomplishing task adaptation. This approach facilitates a more effective transfer of CLIP’s knowledge to the downstream TI-ReID task.

Moreover, the CLIP model focuses on constructing instance-level matching between image-text pairs, with its pre-training objective aimed at learning semantic correlations between images and their corresponding texts, thereby emphasizing one-to-one correspondence. In contrast, TI-ReID requires identity-level retrieval based on natural language descriptions, highlighting cross-modal identity consistency modeling. This shift from “semantic matching” to “identity discrimination” directly limits the transferability of CLIP to TI-ReID tasks.
To address this challenge, Yan et al. \cite{yan2023learning} construct a multi-image multi-text support set during training, perform semantic enhancement for alignment, and finally distill the model into a lightweight version that requires only a single image or text at inference. Gou et al. \cite{gou2025instance} propose a Structure-level Distribution Guided feature learning method, which introduces more stable category-level feature distributions to calibrate the unstable instance-level distributions. Furthermore, Yan et al. \cite{yan2024prototypical} establish an end-to-end Prototypical Prompting framework (Propot) that simultaneously models instance-level and identity-level matching, ensuring precise text-image alignment while enhancing cross-image/text identity consistency modeling.

Although vision-language pre-trained models exhibit strong semantic understanding and achieve impressive performance in image-text alignment tasks, they often rely on an implicit yet unrealistic assumption: all image-text pairs in the training data are perfectly aligned, clean, and error-free.
In the context of TIReID, where most image-text pairs share high semantic consistency but may contain subtle mismatches in fine-grained details, conventional methods struggle to detect such local misalignments, as shown in Fig. 1, ‘D’,thereby degrading discriminative performance.
To address this, Xu et al. \cite{xu2023mining} explicitly model “mismatched word–region pairs” and incorporate them as negative signals during training, enhancing the model’s sensitivity to imperfect matches.
Building upon this idea, Qin et al. \cite{qin2024noisy} formally define such mismatched correspondences as "Noisy Correspondence (NC)" and propose a Robust Dual-branch Embedding (RDE) framework, which mitigates the impact of noisy alignments on model optimization through a noise-tolerant training paradigm.

In TI-ReID, modality information is often incomplete due to subjective or ambiguous textual descriptions and occluded or low-quality images. To enhance the robustness of models under such incomplete information, existing studies commonly adopt cross-modal semantic inference mechanisms to mitigate alignment difficulties arising from modality incompleteness.
Both Fujii et al. \cite{fujii2023bilma} and Xiang et al. \cite{xiang2023learning} introduce masked modeling strategies over image patches and text tokens, employing bidirectional masking and reconstruction to enhance the model’s robustness against local semantic omissions.
Alternatively, Du et al. \cite{du2025graph} construct a semantic-aware graph structure to guide the available modality in semantically completing the missing one, thereby achieving cross-modal reconstruction and alignment from a graph-structural perspective.

Beyond the above discussions, several recent works explore alternative perspectives to further enhance TIReID performance. For instance, Si et al. \cite{qin2025human} introduce an interactive cross-modal learning framework (ICL), which supports multi-turn interaction between users and multi-modal large language models (MLLM) via visual question answering, enabling dynamic query refinement with a focus on person-specific attributes. In addition, Du et al. \cite{du2024bottom} identifie the model’s over-reliance on color cues when image and textual color information align, and proposes a color-robust learning framework by incorporating color-perturbed images as an auxiliary modality, along with fine-grained alignment at the minimal semantic unit level across image, perturbed image, and text to mitigate color bias and enhance semantic discrimination.

Recent studies extend vision-language adaptation for TI-ReID beyond fixed identity prompts and full-model fine-tuning. HAM \cite{jiang2025modeling} learns prompts that represent diverse human annotation styles and uses them to guide multimodal language models in generating varied pedestrian descriptions. Niu et al. \cite{niu2025chatreid} adopts hierarchical progressive tuning to transfer a vision-language model from pedestrian-attribute understanding to fine-grained retrieval and instruction-driven reasoning. From the perspective of parameter-efficient transfer, DM-Adapter \cite{liu2025dm} inserts a sparse mixture of domain-aware adapters into the frozen visual and textual branches of CLIP, allowing different experts to capture complementary aspects of fine-grained person knowledge. Compared with full fine-tuning, prompt and adapter strategies update fewer task-specific parameters and can better preserve pretrained knowledge; however, their efficiency and retrieval value depend on whether the adapted components capture identity-level details rather than only generic pedestrian semantics.

\begin{table}[t]
\centering
\setlength{\tabcolsep}{4pt} 
\footnotesize
\caption{Performance comparison of TI ReID methods on CUHK-PEDES, ICFG-PEDES, and RSTPReid datasets in terms of Rank-1 accuracy and mAP.}
\setlength{\tabcolsep}{4pt} 
\begin{tabular}{c|c|c|cc|cc|cc}
\toprule
\multirow{2}{*}{ } & \multirow{2}{*}{Methods}& \multirow{2}{*}{Venue} & \multicolumn{2}{c|}{CUHK-PEDES} & \multicolumn{2}{c|}{ICFG-PEDES}& \multicolumn{2}{c}{RSTPReid} \\
\cmidrule{4-9}
 & & & R-1 & mAP & R-1 & mAP & R-1 & mAP \\
\hline
\multirow{11}{*}{w/o CLIP}
& LBUL \cite{LBUL} & MM-22 & - & - & - & - & 45.55 & -\\
& ACSA \cite{ACSA} & TMM-23 & - & - & - & - & 48.40 & -\\
& C$_2$A$_2$ \cite{C2A2} & MM-22 & 64.82 & - & - & - & 51.55 & -\\
& FedSH \cite{FedSH} & TMM-23 & 60.87 & - & 55.01 & - & - & -\\
& PBSL \cite{PBSL} & MM-23 & 65.32 & - & 57.84 & - & 47.80 & -\\
& BEAT \cite{Beat} & MM-23 & 65.61 & - & 58.25 & - & 48.10 & -\\
& MANet \cite{MANet} & TNNLS-23 & 65.64 & - & 59.44 & - & - & -\\
& ASAMN \cite{ASAMN} & TIP-23 & 65.66 & - & 57.09 & - & - & -\\
& LCR$^2$S \cite{yan2023learning} & MM-23 & 67.36 & 59.24 & 57.93 & 38.21 & 54.95 & 40.92\\
& TransTPS \cite{TransTPS} & TMM-23 & 68.23 & - & - & - & 56.05 & -\\
& MGCN \cite{MGCN} & TMM-23 & 69.40 & 62.49 & 60.20 & 37.56 & 52.95 & 41.74\\
\hline
\multirow{18}{*}{w/ CLIP}
& CFine \cite{yan2023clip} & TIP-23 & 69.57 & - & 60.83 & - & 50.55 & -\\
& VLP-TPS \cite{VLP-TPS} & arXiv-23 & 70.16 & 66.32 & 60.64 & 42.78 & 50.65 & 43.11\\
& VGSG \cite{VGSG} & TIP-23 & 71.38 & - & 63.05 & - & - & - \\
& IRRA \cite{IRRA} & CVPR-23 & 73.38 & 66.13 & 63.46 & 38.06 & 60.20 & 47.17 \\
& TCB \cite{TCB} & MM-23 & 74.45 & 64.12 & 61.60 & 44.31 & - & -\\
& DCEL \cite{DCEL} & MM-23 & 75.02 & - & 64.88 & - & 61.35 & -\\
& SAL \cite{SAL} & MMM-24 & 69.14 & - & 62.77 & - & - & -\\
& EESSO \cite{EESSO} & IVC-24 & 69.57 & - & 60.84 & - & 53.15 & -\\
& PD \cite{li2024prompt} & arXiv-24 & 71.59 & 65.03 & 60.93 & 36.44 & 56.65 & 45.27\\
& CFAM \cite{CFAM} & CVPR-24 & 72.87 & 64.92 & 62.17 & 39.42 & 59.40 & 49.50 \\
& MACF \cite{MACF} & IJCV-24 & 73.33 & - & 62.95 & -\\
& TBPS-CLIP \cite{cao2024empirical} & AAAI-24 & 73.54 & 65.38 & 65.50 & 39.83 & 61.95 & 48.26\\
& UMSA \cite{UMSA} & AAAI-24 & 74.25 & 66.15& - & - & - & - \\
& LSPM \cite{LSPM} & TMM-24 & 74.38 & 67.74 & 64.40 & 42.60& - & -\\
& IRLT \cite{IRLT} & AAAI-24 & 74.46 & - & 64.72 & - & 61.49 & -\\
& MDRL \cite{MDRL} & AAAI-24 & 74.56 & - & - & - & - & -\\
& FSRL \cite{FSRL} & ICMR-24 & 74.65 & 67.49 & 64.01 & 39.64 & 60.20 & 47.38\\
& Propot \cite{yan2024prototypical}  & MM-24  & 74.89 & 67.12 & 65.12 & 42.93 & 61.87 & 47.82\\
\bottomrule
\end{tabular}
\label{tab:ti-reid}
\end{table}

\subsection{Sketch ReID}

In real-world security and criminal investigation scenarios, acquiring high-quality photographs of suspects or witnesses is often extremely difficult. In many cases, investigators have to rely on sketches drawn from eyewitness memory, which gives rise to the Sketch-ReID task that aims to enable cross-modal identity retrieval between sketches and photographs.

\subsubsection{Datasets}

PKU-Sketch \cite{pang2018cross} is a prominent dataset for sketch-based person re-identification, encompassing 200 identities, each with one hand-drawn sketch and two photographs. The photographs, captured in daylight from two cross-view cameras, are manually cropped to isolate individual subjects. Sketches, created by five professional artists, were iteratively refined from eyewitness descriptions, resulting in diverse drawing styles with an imbalanced distribution. The dataset adheres to a structured protocol that designates 3/4 of the identities per style for training purposes and the remaining 1/4 for testing, resulting in a 150/50 identity split. Each sketch and photograph within the dataset is uniquely annotated, ensuring clarity and precision in matching. This organization mirrors real-world scenarios encountered in surveillance and forensic sketch-photo matching, making PKU-Sketch a crucial benchmark for assessing the efficacy of sketch-based Re-identification (ReID) methods. For additional information and insights, please refer to Table \ref{tab:sketch-data}.

CUFSF database (CUHK Face Sketch FERET) \cite{2011Coupled} is constructed by the Chinese University of Hong Kong based on the FERET dataset, primarily for face sketch synthesis and recognition research. It contains 1,194 identities, each with a face photo under lighting variations and a corresponding sketch. The sketches are drawn by professional artists while viewing the photos, featuring certain shape exaggerations to realistically capture the modality gap between sketches and photos.Detailed information regarding the datasets is provided in Table \ref{tab:sketch-data}.

    



\begin{table}[htbp]
\centering
\footnotesize
\caption{Statistics of Sketch ReID related datasets, including the number of identities, RGB images, sketch images, sketch styles, and training/testing splits.}
\begin{tabular}{cccccc}
\toprule
Dataset & Identities & RGB imgs & Sketch imgs & Style & Training/Testing \\
\hline
\multirow{6}{*}{PKU-Sketch \cite{pang2018cross}} 
 & \multirow{6}{*}{200} 
 & \multirow{6}{*}{400} 
 & \multirow{6}{*}{200} 
 & a (46) & 34/12 \\
 &  &  &  & b (20) & 15/5 \\
 &  &  &  & c (79) & 60/19 \\
 &  &  &  & d (33) & 25/8 \\
 &  &  &  & e (22) & 16/6 \\
 &  &  &  & total (200) & 150/50 \\
\hline
CUFSF \cite{2011Coupled} & 1,194 & 1,194 & 1,194 & - & -\\
\bottomrule
\label{tab:sketch-data}
\vspace{-5mm}
\end{tabular}
\end{table}



\textbf{Summary:}Overall, these sketch-based benchmarks exhibit distinct acquisition and annotation biases. PKU-Sketch is designed for person retrieval under surveillance scenarios but is limited by its relatively small scale and style imbalance among hand-drawn sketches, whereas CUFSF focuses on face sketch-photo matching with sketches generated under controlled conditions. Consequently, existing sketch-based benchmarks only partially capture the diversity and complexity of real-world sketch-based ReID applications.


\subsubsection{Methods}

Sketches are inherently abstract, preserving only structural contours and sparse texture details, while discarding critical visual cues such as color, illumination, and background context. Furthermore, sketches are typically drawn from standard frontal views, whereas real images often exhibit large variations in viewpoint, pose, occlusion, and imaging conditions. Additional semantic divergence arises from the subjectivity of eyewitness descriptions and stylistic differences in sketch rendering.
These factors collectively exacerbate the modality gap between sketches and photos, making cross-modal alignment and identity discrimination particularly challenging. As a compelling extension of traditional person Re-ID, it offers substantial real-world relevance and has garnered increasing attention within the broader field of cross-modal visual understanding. We compare the performance of existing methods in Table \ref{tab:sketch}.

Pang et al. \cite{pang2018cross} first formally introduce the Sketch Re-ID task and propose a cross-domain adversarial feature learning framework consisting of a domain discriminator and two modality-specific generators. Sketches and photos are encoded by separate generators and then fed into the discriminator, which enforces modality confusion through adversarial learning. Meanwhile, identity classifiers in both modalities provide supervision to ensure discriminative identity representations. This approach effectively filters out low-level interference (e.g., texture and color) while preserving high-level semantic information. Meanwhile, Pang et al. release the PKU-Sketch dataset, which plays an important role in driving subsequent studies on Sketch Re-ID.
Li et al. \cite{zhang2022cross} introduce a Cross-Compatible Embedding (CCE) framework combined with a Semantic Consistent Feature Construction (SCFC) scheme, which enables local-level cross-modal feature exchange in a Transformer and constructs semantically consistent global representations to alleviate the modality gap in Sketch Re-ID.
Gui et al. \cite{gui2020learning} propose a multi-level domain-invariant feature learning framework for sketch-based ReID with a staged training strategy. It model first extracts shared global identity features, then incorporates an adversarial domain discriminator to reduce modality discrepancies at the mid-level, and finally applies attention mechanisms to align fine-grained local features across modalities.
To mitigate the inherent information asymmetry between sketches and photos, Chen et al. \cite{chen2023sketchtrans} first generate an auxiliary sketch modality from the photo modality, then employ an asymmetric disentanglement mechanism to decompose photo features into sketch-relevant and sketch-irrelevant parts, transferring the latter to the sketch modality to compensate for missing information, and further introduce a prototype contrastive learning scheme to capture shared modality information.
Liu et al. \cite{liu2024differentiable} introduce an intermediate modality that integrates photo appearance cues with sketch structural abstractions to bridge the modality gap, and further design a tri-modal collaborative learning strategy across sketches, the intermediate modality, and photos to enhance semantic consistency and discriminative capability under limited sketch information.
In Sketch Re-ID, sketches are often influenced by eyewitness memory bias and stylistic variations, leading to significant semantic inconsistencies with real photos. To address this subjectivity issue, Lin et al. \cite{lin2023beyond} construct a large-scale, multi-style sketch-photo dataset to systematically model and analyze the challenges in sketch-based re-identification. The dataset contains 4,763 sketches and 32,668 photos, with each identity represented by multiple sketches drawn by different eyewitnesses and artists, thereby realistically capturing perceptual discrepancies and stylistic diversity. Based on this dataset, the authors introduce two modules to mitigate subjectivity: the NL module, which aggregates multiple sketches of the same identity to counteract discrepancies caused by different eyewitnesses and drawing styles; and the AttrAlign module, which leverages attributes such as gender and clothing as references to facilitate cross-modal alignment between sketches and photos. Recently, Hu et al.~\cite{hu2025cross} propose adaptive incremental prompt tuning to progressively transfer drawing-style knowledge from generic sketch datasets to Sketch-ReID, addressing the limited availability of identity-labeled sketch data while improving domain adaptation to sketch-specific representations.
  Gong et al.~\cite{gong2026theory} propose KTCAA, a few-shot cross-modal framework that combines alignment augmentation with a knowledge transfer catalyst in a meta-learning paradigm, aiming to reduce domain discrepancy and improve robustness to modality perturbations with limited labeled sketches. These studies extend Sketch-ReID beyond conventional feature alignment by considering efficient style adaptation and few-shot cross-modal generalization.

\begin{table*}[htbp]
  \caption{Comparison of Sketch ReID methods in terms of Rank-1, Rank-5, Rank-10 accuracy, and mAP.}
  \label{tab:sketch}
  \footnotesize
  
  \begin{tabular*}{\textwidth}{@{\extracolsep{\fill}} c c c c c c @{}}
    \toprule
    Methods & Venue &  R-1 & R-5  & R-10 & mAP \\
    \midrule 
    TripletSN \cite{yu2016sketch}& CVPR-16 & 9.0 & 26.8 & 42.2 & - \\
    GNSiamese \cite{sangkloy2016sketchy} & TOG-16 & 28.9 & 54.0 & 62.4 & - \\
    AFLNet \cite{pang2018cross}& MM-18 & 34.0 & 56.3 & 72.5 & - \\
    LMDI \cite{gui2020learning}& Neurocomputing-20 & 49.0  & 70.4 & 80.2 & - \\
    CDAC \cite{zhu2022cross}& TIFS-22  & 60.8 & 80.6 & 88.8 & - \\    
    CSIG \cite{chen2021a-cross}& IFTC-21 & 77.6 & 93.0 & 97.0 & - \\
    SketchTrans \cite{chen2022sketch}& MM-22 & 84.6 & 94.8 & 98.2 & - \\
    CCSC \cite{zhang2022cross}& MM-22 & 86.0 & 98.0 & 100.0 & 83.7\\   
    DALNet \cite{liu2024differentiable}& AAAI-24 & 90.0 & 98.6 & 100.0 & 86.2 \\
    \bottomrule
  \end{tabular*}
\end{table*}

\subsection{NLOS ReID}
In recent years, Non-Line-of-Sight Re-identification (NLOS ReID) has emerged as a significant research area that integrates wireless signal perception with computer vision, addressing the limitations of traditional visual sensors for pedestrian identification in occluded or non-line-of-sight scenarios. This approach utilizes various wireless signal technologies, such as radar, Wi-Fi, and ultrasonic systems, to enhance identification capabilities.

 Non-Line-of-Sight ReID extends cross-modal identity retrieval beyond conventional visual observations by exploiting sensing signals that remain informative under occlusion, poor illumination, or limited camera visibility. Existing studies provide important foundations for this direction by investigating whether non-visual observations contain stable identity evidence.

   Guo et al. \cite{guo2024lidar} introduce ReID3D, the first LiDAR-based person re-identification framework, together with the LReID
  dataset and the synthetic LReID-sync dataset. By exploiting geometric features and pre-training for point-cloud completion and shape learning, their work demonstrates that sparse 3D observations can contain identity-relevant structural cues beyond conventional image appearance. However, the sparsity and viewpoint dependence of point clouds make it difficult to establish fine-grained correspondence with dense RGB images, while the synthetic-to-real gap may further affect the transferability of the learned representations.
  Beyond static geometric observations, Liu et al.\cite{liu2024mission} investigate cross-modal ReID using millimeter-wave (mmWave) radar and RGB cameras. Their method aligns radar and image features in a shared representation space and introduces 3D pose estimation as an auxiliary task. This design highlights the value of radar for capturing motion- and geometry-related identity cues under occlusion or poor illumination. It is worth noting that radar measurements are strongly affected by signal reflection, body orientation, multipath propagation, and temporal misalignment, making direct pixel-level correspondence with RGB images inappropriate. Cascio et al.\cite{cascio2025benchmark}introduce Wi-PER81, the first public benchmark for radio-signal-image-based ReID, containing 162,000 wireless signal packets from 81 individuals across two sessions, together with a Siamese baseline for processing signal-magnitude heatmaps. The dataset and baseline establish an important foundation for evaluating identity cues in wireless signals. A key challenge is that the observed Wi-Fi signal patterns are highly dependent on the sensing environment. Variations in devices, rooms, propagation conditions, and body–sensor configurations can substantially alter the measured signals, making stable identity representation considerably more difficult than in vision-based ReID.
Sheng et al. \cite{sheng2025toward} explored person ReID using mmWave radar by leveraging sparse 3D point clouds instead of visual appearance. Their framework combines an Anchor-based Orientation Descriptor (AOD) with a Multi-scale Gait Catch (MGC) module to jointly model local geometric structures and multi-scale gait dynamics for robust identity representation.Zhao et al. \cite{fan2020learning} introduced RF-ReID, which utilizes radio frequency (RF) reflections to extract intrinsic body characteristics rather than appearance cues, enabling long-term, privacy-preserving person re-identification that is robust to clothing changes and occlusions. More recently, Cascio et al. \cite{cascio2025benchmark} further advanced this direction by introducing the Wi-PER81 benchmark, the first public Wi-Fi signal image-based ReID dataset, together with a Siamese baseline for learning discriminative representations from wireless signal heatmaps.

Beyond sparse radar point clouds, recent mmWave ReID studies explore temporal
and image-like signal representations. Han et al.~\cite{han2025mmreid} extracts multi-frequency micro-Doppler patterns to characterize gait dynamics, whereas Wang et al.~\cite{wang2025user} constructs radio images from mmWave echoes to retain
more spatial structure. These approaches demonstrate that radar identity cues
can be represented as geometry, motion spectra, or radio images. Their alignment
with RGB data nevertheless remains nontrivial: micro-Doppler features require
gait-phase and sequence-level correspondence rather than pixel matching, while
radio images differ from optical images in resolution, reflection mechanism,
and sensitivity to body orientation and multipath. Mission~\cite{liu2024mission} directly addresses the cross-modal case by matching a
radar query to an RGB gallery, but broader cross-device and cross-environment
validation remains necessary.

Acoustic sensing provides a related but less mature direction for identity
perception. SonicID~\cite{li2024sonicid} uses active ultrasonic reflections for
near-field user authentication, while acoustic footstep studies
~\cite{wu2023advanced} identify people from gait-related sound patterns.
These results indicate that echoes and footsteps can encode identity-related
shape or motion information, but they should not be regarded as established
acoustic-to-visual ReID methods. Footstep signatures vary with footwear, floor
material, reverberation, background noise, and microphone placement, and their
temporal gait cues have little direct correspondence with appearance-rich RGB
images. Establishing this direction therefore requires synchronized
acoustic--visual datasets, identity-level query--gallery protocols, and
cross-room and cross-device evaluation.


 Despite recent progress, cross-modal alignment between non-visual observations and dense visual imagery remains the primary challenge in NLOS ReID. Unlike RGB images, non-visual modalities provide sparse or indirect observations with fundamentally different representations, making identity correspondence difficult to establish. Learning modality-invariant yet identity-discriminative representations therefore remains a key direction for future research.

\section{Tri-spectral Person Re-identification}

Most existing studies on cross-modal visible–infrared person re-identification (ReID) focus on dual-spectral settings, which alleviate some limitations of single-modal approaches but do not fully address the diverse requirements of extreme conditions. This shortcoming drives the investigation of tri-spectral ReID, which utilizes complementary information from RGB, NIR, and TIR modalities to better meet the challenges of complex surveillance environments.
RGB images capture visible light, offering rich color and texture information but are highly sensitive to illumination variations. Infrared images can be classified by their imaging principles: NIR images are generated from near-infrared reflection, providing stable structural data under low light but suffering detail loss in bright conditions; TIR images, derived from heat radiation, facilitate effective human–background separation in darkness or adverse weather, though they lack fine-grained appearance details. Leveraging these complementary properties, tri-spectral person re-identification research aims to integrate RGB, NIR, and TIR data to enhance the robustness and discriminative power of identity representations.

The selection of RGB, near-infrared (NIR), and thermal-infrared (TIR) as the primary tri-spectral configuration is motivated by their complementary sensing characteristics under varying illumination conditions. RGB provides rich appearance details, NIR preserves structural information under challenging lighting, and TIR captures thermal patterns in dark environments. Therefore, existing tri-spectral ReID studies combine these modalities to exploit complementary visual cues across diverse scenarios. Nevertheless, this configuration represents a practical choice supported by existing sensing systems and benchmarks rather than a universally optimal solution.

\subsection{Datasets}
RGBNT201~\cite{zheng2021robust} is a benchmark dataset designed for
  tri-spectral person re-identification. It contains synchronized RGB, NIR, and
  TIR observations of 201 identities captured by four cameras, providing an
  experimental basis for studying spectral complementarity under challenging
  imaging conditions. Nevertheless, its relatively limited identity and camera
  coverage constrains the diversity of populations, scenes, and sensing
  conditions represented in the benchmark. Moreover, the use of synchronized
  and complete modality triplets favors fixed three-stream fusion, whereas
  practical systems may encounter imperfect registration or unavailable
  modalities. Performance on RGBNT201 should therefore be interpreted as evidence
  of fusion effectiveness under its acquisition protocol, rather than as a
  complete measure of robustness across different sensors and deployment
  environments.


\subsection{Methods}


Wang et al. \cite{wang2024top} propose TOP-ReID, a novel multi-spectral ReID framework, which performs token-level permutation to achieve feature alignment and aggregation across spectra, thereby enhancing discriminative feature interaction while preserving modality-specific details. To address the issue of missing spectral information in real-world scenarios, a token-level reconstruction constraint is further introduced to effectively reduce distribution gaps among different spectra.
In a subsequent work, Wang et al. \cite{wang2025decoupled}  further introduce the DeMo framework to address the challenges of dynamic imaging quality variations and the weakening of modality-specific information in multi-modal object re-identification. The framework employs a Hierarchical Decoupling Module (HDM) to separate modality-specific and shared information, and introduces an Attention-Triggered Mixture of Experts (ATMoE) mechanism to perform dynamic weight allocation.

\begin{table*}[htbp]
  \footnotesize
  \caption{Comparison of tri-spectral ReID methods on the RGBNT201 dataset is conducted, evaluating metrics such as mean Average Precision (mAP), Rank-1, Rank-5, and Rank-10 accuracy.}
  \label{tab:tri-spectral} 
  
  \begin{tabular*}{\textwidth}{@{\extracolsep{\fill}} c c c c c c @{}}
    \toprule
    Methods & Venue & mAP & R-1 & R-5  & R-10  \\
    \midrule 
    HAMNet \cite{li2020multi}&AAAI-20 & 27.7 & 26.3 & 41.5 & 51.7 \\
    PFNet \cite{zheng2021robust}&AAAI-21 & 38.5 & 38.9 & 52.0 & 58.4 \\
    IEEE \cite{wang2022interact}&AAAI-22  & 47.5 & 44.4 & 57.1 & 63.6  \\
    DENet \cite{zheng2023dynamic}&arXiv-23  & 42.4 & 42.2 & 55.3 & 64.5  \\
    UniCat \cite{crawford2023unicat}&arXiv-23  & 57.0 & 55.7 & - & - \\
    HTT \cite{wang2024heterogeneous}&AAAI-24 & 71.1 & 73.4 & 83.1 & 87.3 \\
    TOP-ReID \cite{wang2024top}&AAAI-24  & 72.3 & 76.6 & 84.7 & 89.4 \\   
    EDITOR \cite{zhang2024magic}&CVPR-24  & 66.5 & 68.3 & 81.1 & 88.2 \\
    RscNet \cite{yu2024representation}&TCSVT-24  &  68.2 & 72.5 & - & - \\
    LRMM \cite{wu2025lrmm}& ESWA-25  & 52.3 & 53.4 & 64.6 & 73.2 \\    
    DeMo \cite{wang2025decoupled}& AAAI-25  & 73.7 & 80.5 & 88.3 & 91.5  \\
    \bottomrule
  \end{tabular*}
\end{table*}

\section{Multi-modal Person Re-identification}

Cross-modal person re-identification continuously drives the field toward greater adaptability and practical utility across diverse scenarios, significantly enhancing retrieval performance under challenging conditions such as complex illumination and missing visual information. Nevertheless, existing research and datasets are predominantly limited to specific modality combinations, whereas real-world applications require ReID systems capable of accurately identifying individuals across both single-modality and multi-modality inputs.


Building on this practical objective, research has gradually expanded from binary cross-modal settings to multi-modal configurations, enabling more flexible inputs and more comprehensive information integration.  Its primary objective is to integrate complementary identity evidence across modalities while controlling cross-source redundancy and accommodating variations in modality availability. Although effective multi-modal fusion still requires
  cross-modal alignment, its principal emphasis lies in combining complementary
  information rather than performing retrieval between a fixed pair of
  heterogeneous query and gallery domains. Existing multi-modal ReID frameworks mainly encompass two categories: tri-modal settings (image, text, and sketch) and five-modal settings (visible image, infrared image, text, sketch, and color sketch). In this chapter, we systematically review related methods and datasets, summarizing their technical characteristics and application scenarios.

\subsection{Datasets}
Zhai et al. \cite{zhai2022trireid} construct a multi-modal ReID dataset that encompasses text, sketch, and RGB images by extending the existing PKU-SketchReID dataset, where an image captioning model is employed to automatically generate textual descriptions for the RGB images.
Chen et al. \cite{chen2023towards} propose and construct three large-scale multi-modal person re-identification datasets, namely Tri-CUHK-PEDES, Tri-ICFG-PEDES, and Tri-RSTPReid, to address the lack of multi-modal benchmarks in existing research. These datasets are extended from the text-based datasets CUHK-PEDES\cite{li2017person}, ICFG-PEDES\cite{ding2107semantically}, and RSTPReid\cite{zhu2021dssl}, respectively, by introducing the sketch modality, thereby forming a tri-modal structure that includes RGB images, textual descriptions, and synthesized sketches.
Zuo et al. \cite{zuo2025reid5o} propose the ORBench dataset, which is constructed on the basis of existing visible–infrared benchmarks SYSU-MM01 \cite{Wu_2017_ICCV} and LLCM \cite{DEEN}. Specifically, representative pedestrian samples are first selected from RGB and infrared images, while discarding those with poor imaging quality or weak clothing representativeness. Then, an online image generation model \cite{doubao} combined with manual supervision is introduced to produce color pencil drawings for each pedestrian, which are further transformed into sketch-style images using the Meitu software \cite{meitu}. In addition, human annotators are employed to provide detailed textual descriptions for each RGB sample, thereby enriching the dataset with semantic information. In total, ORBench contains 1,000 identities with 45,113 RGB images, 26,071 infrared images, 18,000 color pencil drawings, 18,000 sketches, and 45,113 textual descriptions. Table \ref{tab:multi-data} summarizes the detailed statistics of the datasets.


\subsection{Methods}
As two representative descriptive modalities, text and sketch each possess unique advantages. Text, despite its abstract nature, can convey rich semantic information such as body pose, clothing details, and carried items. Sketches, on the other hand, while belonging to the visual modality, excel at depicting contours, structural shapes, and local textures, but often lack color information and exhibit limitations in conveying fine-grained details and complete semantic content. Therefore, text and sketch are inherently complementary in terms of semantic representation. Effectively integrating these two descriptive modalities holds the potential to compensate for the weaknesses of each individual source, enabling the construction of more comprehensive and discriminative person representations, and offering valuable support for cross-modal retrieval in complex real-world scenarios.
Zhai et al. \cite{zhai2022trireid} are the first to use both text and sketch as query modalities for multi-modal person re-identification. Their method extracts features separately for textual and sketch inputs, and then employs an attention-based fusion mechanism to generate a comprehensive semantic representation. To further enhance cross-modal matching, they introduce a generative adversarial learning framework, where multiple loss functions progressively align descriptive and image features. Subsequent studies point out that independently pre-training text and image modalities may limit generalization capability. Moreover, in real-world scenarios, the availability of both text and sketch modalities is not always guaranteed, which highlights the need for practical systems capable of handling diverse query modalities with consistent performance. To address this issue, Chen et al. \cite{chen2023towards} propose the UNReID framework, which unifies Text-to-RGB, Sketch-to-RGB, and Text+Sketch-to-RGB retrieval into a single paradigm. This framework supports both cross-modal and multi-modal retrieval, and the authors further construct three multi-modal datasets—Tri-CUHK-PEDES, Tri-ICFG-PEDES, and Tri-RSTPReid—based on existing text–image datasets. Ha et al.~\cite{ha2025multi} introduce MP-ReID, which contains RGB, infrared, and thermal observations acquired from ground and aerial platforms, together with Uni-Prompt ReID. The method combines
  modality-aware and platform-aware prompts with visual context to condition a shared representation on different sensing and acquisition settings. These studies illustrate two complementary forms of modality flexibility: adapting retrieval to different combinations of available queries and adapting a unified model to heterogeneous sensors and  platforms.



With the continuous advancement of research, multi-modal person ReID has been further extended to a five-modality setting. Zuo et al. \cite{zuo2025reid5o} introduce infrared images and color pencil drawings as additional query modalities on top of RGB, text, and sketch, and propose a unified multi-modal person re-identification framework named ReID5o. This model not only supports single-modality inputs but also flexibly handles combinations of multiple modalities, thereby significantly enhancing the adaptability and robustness of both cross-modal and multi-modal retrieval. Specifically, ReID5o first employs a multi-modal tokenizing assembler to project heterogeneous inputs into a shared embedding space; then applies a unified encoder to extract modality-shared features, together with a multi-expert router to explicitly capture modality-specific representations; meanwhile, a feature mixture mechanism is adopted to fuse information from multiple sources, and cross-modal alignment strategies are integrated to optimize the training objective.

\begin{table*}
  \caption{Statistics of multimodal person re-identification datasets. The table details the number of identities and images for each modality (e.g., RGB, sketch, text) across several benchmarks.}
  \label{tab:datasets-multi}
  \setlength{\tabcolsep}{5pt}  
  \resizebox{\textwidth}{!}{
  \begin{tabular}{ccccccc}
    \toprule
     Datasets & Identities & RGB Imgs & IR Imgs & Sketch Imgs & Texts & Color Pencil Imgs\\
    \midrule
    TriReID \cite{zhai2022trireid} & 200 & 5,600 & - & 200 & 5,600 & - \\
    Tri-CUHK-PEDES\cite{chen2023towards} & 13,003 & 40,206 & -& 40,206 &  80,440 & -\\
    Tri-ICFG-PEDES\cite{chen2023towards} & 4,120 & 54,522 & - & 54,522 &  54,522 & -\\
    Tri-RSTPReid\cite{chen2023towards} & 4,101 & 20,505 & - & 20,505 &  41,010 & -\\
    ORBench \cite{zuo2025reid5o} & 1,000 & 45,114 & 26,071 & 18,000 & 45,113 & 18,000\\
    
    \bottomrule
    \label{tab:multi-data}
    \vspace{-4mm}
  \end{tabular}
  }
\end{table*}

ReID5o~\cite{zuo2025reid5o} provides a controlled setting for examining modality complementarity because multiple query-modality combinations are
  evaluated using the same model and ORBench protocol. For example, taking the text-only query as a reference, its combination with infrared, color-pencil, or sketch information increases mAP from 54.88\% to 70.77\%, 81.14\%, and 74.01\%, respectively. This comparison does not imply that ReID5o relies on a fixed text-centered modality configuration; rather, it provides one illustrative subset of its flexible query combinations for measuring the incremental contributions of different modalities.
  As summarized in Table~\ref{tab:modality-complementarity}, these contributions are not uniform. Adding infrared to the text+color-pencil query improves mAP by 5.61 percentage points, whereas adding sketch to the same pair improves it
  by only 1.91 points. Moreover, the full text+infrared+color-pencil+sketch query reaches 87.46\% mAP, only 0.71 points above text+infrared+color-pencil. These controlled comparisons provide task-level evidence of modality complementarity while also revealing redundancy
  and diminishing returns. Therefore, increasing the number of modalities does not necessarily constitute the best deployment choice when sensing cost, input quality, and modality availability are jointly considered.

\begin{table}[htbp]
\centering
\color{black}
\footnotesize
\setlength{\tabcolsep}{30pt}        
\renewcommand{\arraystretch}{1.15}

\caption{Quantitative analysis of modality complementarity under the unified
ReID5o setting on ORBench. T, I, C, and S denote text, infrared image,
color-pencil image, and sketch, respectively. $\Delta$mAP is computed against
the indicated lower-order reference combination.}

\label{tab:modality-complementarity}

\begin{tabular}{lccc}
\toprule
\textbf{Query modalities} & \textbf{Rank-1} & \textbf{mAP} & \textbf{$\Delta$mAP} \\
\midrule

\multicolumn{4}{l}{\textit{Single-modality reference}} \\
T                         & 63.15 & 54.88 & -- \\

\addlinespace

\multicolumn{4}{l}{\textit{Two-modality combinations}} \\
T+I                       & 81.19 & 70.77 & +15.89 \\
T+C                       & 91.01 & 81.14 & +26.26 \\
T+S                       & 84.71 & 74.01 & +19.13 \\

\addlinespace

\multicolumn{4}{l}{\textit{Three-modality combinations}} \\
T+I+C                     & 95.85 & 86.75 & +5.61 \\
T+I+S                     & 92.85 & 81.98 & +7.97 \\
T+C+S                     & 93.19 & 83.05 & +1.91 \\

\addlinespace

\multicolumn{4}{l}{\textit{All modalities}} \\
T+I+C+S                   & 96.79 & 87.46 & +0.71 \\

\bottomrule
\end{tabular}

\end{table}

\section{Proposed baseline}
 

Drawing on the comprehensive review presented in this survey, we observe a clear evolution of VI-ReID research from early CNN-based architectures toward Transformer-based paradigms. While CNN-based methods have achieved remarkable success in feature extraction and modality alignment, they primarily rely on local texture cues and often suffer performance degradation when processing infrared images that lack color and fine-grained appearance details. This observation motivates us to adopt a Transformer-based baseline as a representative modern architecture for validating the design principles summarized throughout the survey. Compared with CNNs, Transformers are better suited for modeling global structural and shape information, providing stronger representational capacity under the infrared modality and facilitating the learning of modality-invariant identity features. Moreover, their patch-token representation naturally enables fine-grained cross-modal semantic interactions, allowing more effective local feature alignment and improved discriminative capability.

Recent advancements in Transformer architectures have spurred significant progress in person re-identification. Various studies have sought to exploit the structural advantages of Transformers to overcome the limitations of traditional CNNs. For instance, Li et al. \cite{DCformer} introduce multiple class tokens with self-diverse constraints to generate compact and diverse embedding subspaces, enhancing discrimination for similar identities. Zhu et al. \cite{AAformer} propose an auto-alignment mechanism that aligns patch-level features to mitigate the impact of misalignment and occlusion. Tan et al. \cite{partformer} further explore fine-grained representation by disentangling attention heads via a Head Disentangling Block and enforcing diversity constraints. Li et al. \cite{pyramidal} incorporate a pyramid architecture with convolutional patch embedding and auxiliary embedding to model multi-scale and non-visual biases



These advances not only demonstrate the potential of Transformers in ReID tasks but also provide new directions for further exploration. Inspired by this, we propose a unified baseline for visible-infrared person re-identification (VI-ReID) that supports both supervised and unsupervised learning paradigms. The architecture is built upon a dual-path Transformer, where visible and infrared images are processed through separate modality-specific branches to extract shallow appearance features. Additionally, we incorporate the random channel augmentation strategy introduced by \cite{yeCA} into the visible stream to enhance joint training. This dual-path design enables the model to capture modality-specific representations and leverage the complementary nature of visible and infrared inputs. It is particularly effective under challenging conditions such as blur, low resolution, occlusion, illumination changes, and viewpoint variations, enhancing the system’s robustness and accuracy.

 \begin{figure*}[t]
        \centering
        \includegraphics[width=1\linewidth]{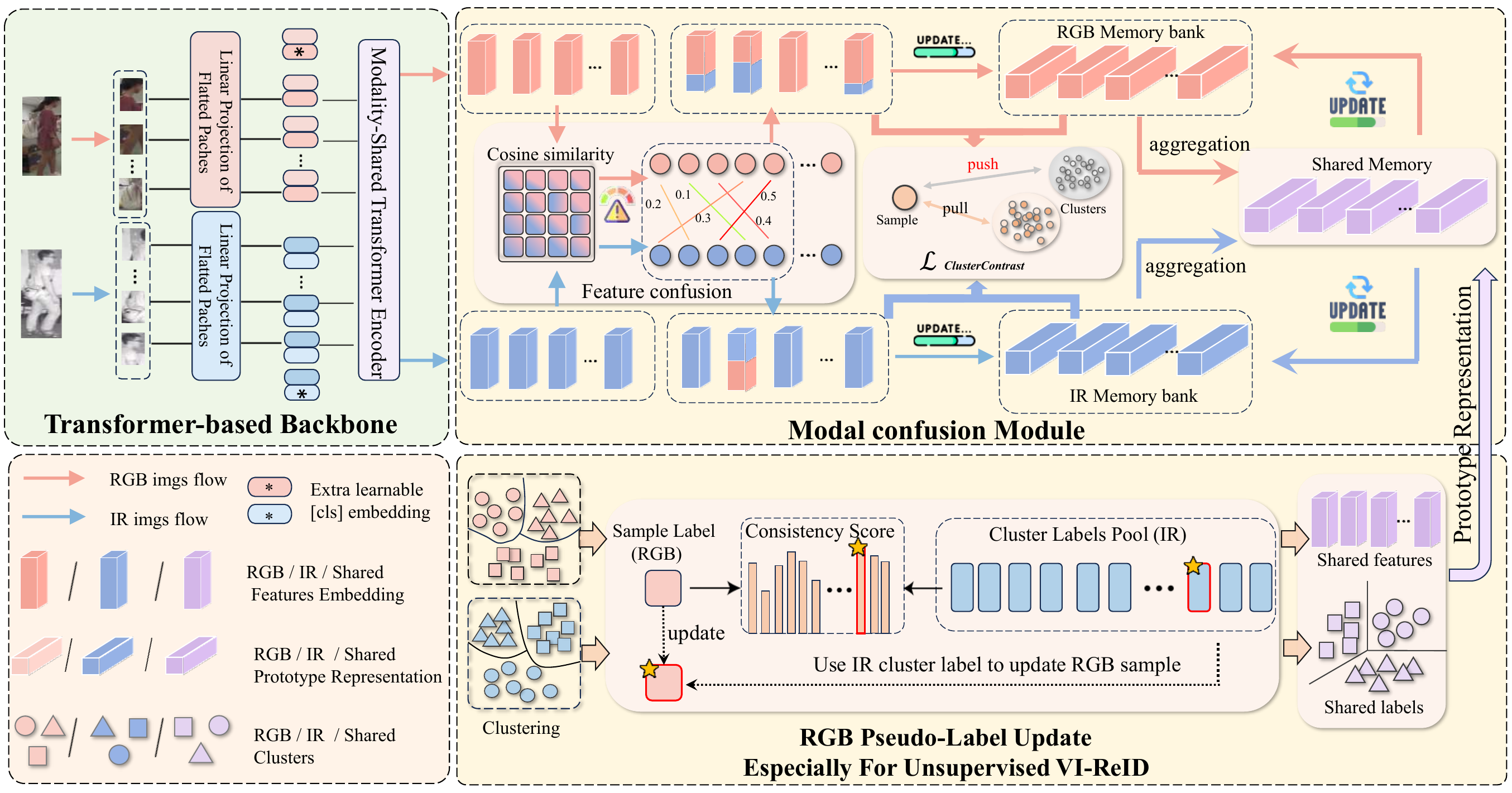}
        \caption{Overall framework of our method. A modality-shared Transformer backbone extracts features from RGB and IR inputs. We introduce a cluster contrast loss to strengthen representation learning and implement a cross-modal label association module that propagates IR cluster labels to RGB samples, thereby facilitating cross-modal label alignment.}
    \label{fig:baseline}
    \vspace{-3mm}
    \end{figure*}

On top of the dual branches, a shared Transformer encoder is used to learn modality-invariant identity embeddings in a unified feature space. 
Inspired by DC-Former \cite{DCformer} , we introduce the class token as a global identity aggregator, which interacts with patch tokens to capture holistic identity information and obtain more discriminative and robust identity representations.
To further bridge the modality gap, we design a simple yet effective modality confusion strategy. Initially, we employ a feature-level weighted confusion mechanism to compel the model to learn modality-invariant representations. Subsequently, we maintain a shared memory bank to integrate multi-modal information. During the optimization phase, we utilize cluster contrastive learning to achieve a more compact clustering distribution in the feature space by pulling samples closer to their assigned cluster centroids while pushing away heterogeneous centroids.
In the context of unsupervised learning, pseudo-labels generated via clustering inevitably contain noise, which detrimentally affects training efficacy. Recognizing that the IR modality often retains more stable structural information under adverse conditions (e.g., illumination variations and color interference), we propose leveraging the IR clustering results to assist in updating the RGB pseudo-labels. Specifically, we propagate the cluster labels from the IR modality to RGB samples based on a cross-modal affinity graph to reconstruct the RGB pseudo-labels, thereby enhancing their accuracy. This strategy effectively mitigates the semantic bias induced by clustering and bolsters the robustness and discriminability of cross-modal pseudo-supervised learning.

The architecture of our proposed baseline is shown schematically in Fig. \ref{fig:baseline}.
Our method demonstrates superior overall performance on both the RegDB and SYSU-MM01 datasets. Under both supervised and unsupervised settings, the results significantly surpass mainstream baseline methods, fully highlighting the potential and competitiveness of the Transformer framework for the VI-ReID task. 
Overall, this baseline provides a strong starting point for Transformer-based VI-ReID and holds considerable value as a unified evaluation platform for subsequent approaches.
The comparison with the current classical baselines on two datasets is shown in Table \ref{tab:results}.

\begin{table*}[t] 
\centering
\caption{Comparison of our proposed method with classical models on the SYSU-MM01 and RegDB datasets.}
\label{tab:results}

\footnotesize 
\setlength{\tabcolsep}{3pt} 

\begin{tabular*}{\textwidth}{@{\extracolsep{\fill}} c l c c c c c c c c c @{}} 
\toprule

\multirow{3}{*}{ } & \multirow{3}{*}{Methods}& \multirow{3}{*}{Venue} & \multicolumn{4}{c}{SYSU-MM01} & \multicolumn{4}{c}{RegDB} \\
\cmidrule(lr){4-7} \cmidrule(lr){8-11} 
 & & & \multicolumn{2}{c}{All Search} & \multicolumn{2}{c}{Indoor Search} & \multicolumn{2}{c}{VIS to IR} & \multicolumn{2}{c}{IR to VIS} \\
\cmidrule(lr){4-5} \cmidrule(lr){6-7} \cmidrule(lr){8-9} \cmidrule(lr){10-11}
 & & & R-1 & mAP & R-1 & mAP & R-1 & mAP & R-1 & mAP \\
\midrule

\multirow{8}{*}{Supervised} 
 & DDAG \cite{ye2020dynamic} & ECCV-20 & 54.75 & 53.02 & 61.02 & 67.98 & 69.34 & 63.46 & 68.06 & 61.80 \\
  & AGW \cite{AGW}   & TPAMI-21 &  47.50 & 47.65 & 54.17 & 62.97 & 70.05 & 66.37 & 70.49 & 65.90 \\
  & CAJ \cite{yeCA}   & ICCV-21 &  69.88 & 66.89 & 76.26 & 80.37 & 85.03 & 79.14 & 84.75 & 77.82\\
 & MCLNet \cite{hao2021cross} & ICCV-21 & 65.40 & 61.98 & 72.56 & 76.58 & 80.31 & 73.07 & 75.93 & 69.49 \\
  & FMCNet \cite{zhang2022fmcnet}  & CVPR-22 &  66.34 & 62.51 & 68.15 & 74.09 & 89.12 & 84.43 & 88.38 & 83.86\\
  & DART \cite{yang2022learning} & CVPR-22 & 68.72 & 66.29 & 72.52 & 78.17 & 83.60 & 75.67 & 81.97 & 73.78 \\
 & DEEN \cite{DEEN} & CVPR-23 & \textbf{74.7} & \textbf{71.8} & \textbf{80.3} & \textbf{83.3} & 91.1 & 85.1 & 89.5& 83.3 \\
  & \textbf{Ours} & - & 69.93 & 68.91 & 76.07 & 81.50 & \textbf{93.48 }& \textbf{88.72} & \textbf{92.61} & \textbf{87.72} \\
  
\midrule 

\multirow{7}{*}{Unsupervised} 
 & OTLA \cite{wang2022OTLA} & ECCV-22 &  29.90 & 27.10 & 29.80 & 38.80 & 32.90 & 29.70 & 32.10 & 28.60 \\ 
 & ADCA \cite{yang2022ADCA} & MM-22 & 45.51 & 41.73 & 50.60 & 59.11 & 67.20 & 64.05 & 68.48 & 63.81 \\
 & PGM \cite{wu2023PGM} & CVPR-23 &  57.27 & 51.78 & 56.23 & 62.74 & 69.48 & 65.41 & 69.85 & 65.17 \\
 & DOTLA \cite{cheng2023DOTLA}& MM-23 & 50.36  & 47.36 & 53.47 & 61.73 & 85.63 & 76.71 & 82.91 & 74.97 \\
 & MBCCM \cite{cheng2023MBCCM}& MM-23 & 53.14 & 48.16 & 55.21 & 61.98 & 83.79 & 77.87 & 82.82 & 76.74 \\
 & UntransReID \cite{ye2025transformer} & IJCV-25 &  51.90 & 52.50 & 57.50 & 66.00 & 76.30 & 69.90 & 76.80 & 69.30 \\
& \textbf{Ours} & - & \textbf{60.87} & \textbf{59.56} & \textbf{66.13} & \textbf{73.21} &\textbf{90.87} & \textbf{85.32} & \textbf{90.67} & \textbf{84.61}   \\
\bottomrule
\end{tabular*}
\end{table*}

\subsection{Implementation Details}
The proposed baseline is implemented in PyTorch using a ViT-Base/16 backbone initialized with pretrained weights. RGB and infrared images are processed by separate patch embedding layers and subsequently passed through a shared Transformer encoder comprising 12 blocks, 12 attention heads, and a 768-dimensional
  embedding space. The output of the shared class token is processed by a Batch
  Normalization neck to obtain the identity representation.
  All images are resized to \(288\times144\). Training augmentation includes
  random cropping, horizontal flipping, color jittering, random erasing, and
  channel augmentation. The model is trained for 50 epochs using SGD with a
  batch size of 256, eight instances per identity, an initial learning rate of
  \(3.5\times10^{-4}\), a momentum of 0.9, and a weight decay of
  \(5\times10^{-4}\). The learning rate is reduced by a factor of 0.1 every
  20 epochs.
  Under supervised training, the objective combines the RGB and infrared
  memory-based contrastive losses with a weighted regularized triplet loss.
  For unsupervised training, RGB and infrared pseudo-labels are generated
  separately using DBSCAN on the re-ranked Jaccard distance. We set the DBSCAN
  threshold to 0.6, the minimum number of samples to 4, and \(k_1=30\) and
  \(k_2=6\). The contrastive temperature and memory momentum are set to 0.05
  and 0.1, respectively. Cross-modal similarities are subsequently used to
  associate infrared clusters with RGB samples and refine the RGB pseudo-labels.

\subsection{Ablation Study Design}

  To further justify the key design choices of the proposed Transformer-based baseline, we conduct an ablation
  study in both supervised and unsupervised VI-ReID settings. The ablation focuses on three components that are central
  to the proposed framework: the modality-shared Transformer encoder, the feature-level modality confusion strategy, and
  the IR-guided RGB pseudo-label refinement module. These components correspond to the main motivations discussed in the
  survey, namely learning modality-invariant identity representations, reducing the heterogeneous modality gap, and
  improving cross-modal label association under noisy pseudo-supervision.

Table~\ref{tab:baseline_ablation} summarizes the ablation protocol. In the supervised setting, we evaluate the
  contributions of the modality-shared Transformer encoder and the feature-level modality confusion strategy. In the
  unsupervised setting, we further evaluate the IR-guided RGB pseudo-label refinement module, while retaining the other
  training components and optimization settings of the proposed baseline. All ablation variants are trained and
  evaluated under the same protocols as the corresponding full model. The experimental results are reported on SYSU-MM01
  and RegDB using Rank-1 accuracy and mAP.

 \begin{table*}[t]
  \centering
  \caption{Ablation study of the proposed Transformer-based VI-ReID baseline. ``MSTE'' denotes the modality-shared
  Transformer encoder, ``FMC'' denotes feature-level modality confusion, and ``IR-PR'' denotes IR-guided RGB pseudo-
  label refinement. All other components and training settings are kept identical to those of the corresponding proposed
  baseline.}
  \label{tab:baseline_ablation}
  \footnotesize
  \setlength{\tabcolsep}{3pt}
  \begin{tabular*}{\textwidth}{@{\extracolsep{\fill}} c l c c c c c c c c c @{}}
  \toprule
  \multirow{2}{*}{Setting} &
  \multirow{2}{*}{Variant} &
  \multirow{2}{*}{MSTE} &
  \multirow{2}{*}{FMC} &
  \multirow{2}{*}{IR-PR} &
  \multicolumn{2}{c}{SYSU-MM01} &
  \multicolumn{2}{c}{RegDB V$\rightarrow$I} &
  \multicolumn{2}{c}{RegDB I$\rightarrow$V} \\
  \cmidrule(lr){6-7}
  \cmidrule(lr){8-9}
  \cmidrule(lr){10-11}
  & & & & & R-1 & mAP & R-1 & mAP & R-1 & mAP \\
  \midrule
  \multirow{3}{*}{Supervised}
  & Proposed baseline & \cmark & \cmark & -- & 69.93 & 68.91 & 93.48 & 88.72 & 92.61 & 87.72 \\
  & w/o MSTE & \xmark & \cmark & -- & 67.42 & 66.58 & 90.96 & 86.14 & 90.18 & 85.27 \\
  & w/o FMC & \cmark & \xmark & -- & 67.21 & 66.86 & 91.66 & 86.31 & 91.03 & 85.78 \\
  \midrule
  \multirow{4}{*}{Unsupervised}
  & Proposed baseline & \cmark & \cmark & \cmark & 60.87 & 59.56 & 90.87 & 85.32 & 90.67 & 84.61 \\
  & w/o MSTE & \xmark & \cmark & \cmark & 57.76 & 56.48 & 87.62 & 82.36 & 87.41 & 81.73 \\
  & w/o FMC & \cmark & \xmark & \cmark & 59.18 & 57.91 & 88.94 & 83.71 & 88.71 & 83.02 \\
  & w/o IR-PR & \cmark & \cmark & \xmark & 58.64 & 57.36 & 88.37 & 82.86 & 88.21 & 82.24 \\
  \bottomrule
  \end{tabular*}
  \end{table*}

This ablation study is designed to clarify the individual and joint contributions of the
  proposed modules. In the supervised setting, we evaluate the effects of the modality-shared
  Transformer encoder (MSTE) and the feature-level modality confusion (FMC) strategy by comparing
  the full model with variants containing neither module or only one of them. MSTE examines
  whether sharing high-level Transformer representations across modalities facilitates visible-
  infrared alignment, while FMC evaluates the benefit of explicitly reducing modality
  discrepancies at the feature level. In the unsupervised setting, we further introduce IR-guided
  RGB pseudo-label refinement (IR-PR). By comparing single-module, pairwise-combination, and full-
  model variants, we investigate whether IR-PR improves the reliability of RGB pseudo-labels and
  whether it complements MSTE and FMC. Overall, the results demonstrate the individual
  contributions and complementary effects of these modules.

\section{Future Research Perspectives}

This section further discusses the future research potential and development directions of VI-ReID, aiming to provide insights for advancing both model performance and practical applicability.

\textbf{Identity-aware Foundation Models}
Vision-language foundation models, such as CLIP, BLIP, and GPT-4o, have demonstrated remarkable capabilities in cross-modal representation learning. However, these models are primarily optimized for semantic alignment rather than identity-level discrimination, which limits their direct applicability to person re-identification. Future research should investigate identity-aware adaptation strategies that preserve fine-grained identity cues while maintaining the strong semantic priors of foundation models. Possible directions include identity-centric prompt learning, modality-aware token interaction, instance-level contrastive objectives, and lightweight parameter-efficient adaptation. Such designs could enable foundation models to better distinguish visually similar individuals while maintaining robust cross-modal generalization.

\textbf{Causal Representation Learning for Modality Disentanglement}
A fundamental challenge in VI-ReID lies in disentangling identity information from modality-specific variations, such as illumination, sensor characteristics, and environmental conditions. Most existing approaches rely on feature alignment or adversarial learning, which often fail to explicitly distinguish causal identity factors from nuisance modality attributes. Future research may benefit from causal representation learning, which aims to separate invariant identity semantics from modality-dependent characteristics by modeling the underlying causal relationships. Such a framework could improve robustness to unseen sensors and environments while enhancing the interpretability and generalization ability of cross-modal ReID systems.

\textbf{Neuro-symbolic Sketch Understanding.}
Sketch-ReID is challenging because sketches are abstract, subjective, and structurally sparse. Existing methods mainly rely on end-to-end feature alignment, which may overlook explicit human body structure, part relations, and attribute logic. Neuro-symbolic learning offers a concrete direction by combining neural representation learning with symbolic reasoning over body parts, attributes, and spatial relations. For example, a model can first extract neural features from sketch contours and photo regions, and then reason over symbolic structures such as ''upper clothing-lower clothing'', ''carried object--body part'', or ''part adjacency''. This can help bridge the gap between subjective sketch strokes and identity-related visual evidence in real photos, especially when color and texture cues are missing.

\textbf{World Models for Long-term Identity Reasoning}
Recent advances in world models have demonstrated impressive capabilities in learning dynamic environments and predicting future observations. Instead of merely generating synthetic training samples, world models could enable VI-ReID systems to reason about identity evolution across viewpoints, temporal changes, and sensor modalities. By modeling long-term pedestrian trajectories, behavioral consistency, and environmental context, world models may facilitate more reliable identity association under severe occlusion, missing observations, or long-term re-identification scenarios. Integrating predictive reasoning with identity representation learning therefore represents a promising research direction beyond conventional image matching.

\textbf{Accuracy-aware Privacy Protection.}
 Current cross-modal person re-identification datasets primarily rely on real-world captured
images, raising concerns regarding identity exposure and privacy protection. As privacy regulations
become increasingly stringent, developing privacy-preserving ReID systems without substantially
compromising retrieval performance has become an important research direction. One promising
direction is to jointly optimize identity discrimination and privacy preservation by learning repre-
sentations that remain highly discriminative for identity matching while minimizing the leakage of
sensitive personal information. For example, representation disentanglement could be employed to
separate retrieval-relevant identity representations from privacy-sensitive visual information, while
differential privacy may be incorporated during feature learning to reduce the memorization of in-
dividual samples without destroying the relative feature structure required for retrieval. In addition,
generative models could be leveraged to synthesize privacy-preserving training samples that retain
identity consistency while concealing sensitive visual attributes. Achieving an effective balance be-
tween privacy protection and identity discrimination will be crucial for the practical deployment of cross-modal ReID systems.

\textbf{Next-Generation ReID Benchmarks
}
Existing benchmarks have significantly advanced person re-identification by enabling standardized evaluation and fair comparison. However, they inevitably contain dataset-specific biases caused by sensing configurations, acquisition environments, and annotation strategies, making high performance not always indicative of real-world robustness. Future benchmarks should emphasize greater diversity in sensing devices, environments, annotation strategies, and modality sources. More comprehensive evaluation protocols, including cross-dataset testing, source-/device-disjoint evaluation, and robustness assessment under missing or degraded modalities, are also needed. These efforts can promote the learning of transferable identity representations beyond benchmark-specific correlations.

\textbf{Identity-aware Foundation Models}
Vision-language foundation models, such as CLIP, BLIP, and GPT-4o, have demonstrated remarkable capabilities in cross-modal representation learning. However, these models are primarily optimized for semantic alignment rather than identity-level discrimination, which limits their direct applicability to person re-identification. Future research should investigate identity-aware adaptation strategies that preserve fine-grained identity cues while maintaining the strong semantic priors of foundation models. Possible directions include identity-centric prompt learning, modality-aware token interaction, instance-level contrastive objectives, and lightweight parameter-efficient adaptation. Such designs could enable foundation models to better distinguish visually similar individuals while maintaining robust cross-modal generalization.

\textbf{Acknowledgements.} This work is partially supported by National Natural Science Foundation of China under Grants (62302351,62501428,62376201,62306215), Postdoctoral Fellowship Program of CPSF under Grant Number ({GZC20241268}, 2024M762479), Hubei Provincial Natural Science Foundation of China (2025AFB219).


\textbf{Conflict of interest statement.} The authors declare that there are no conflict of interests, we do not have any possible conflicts of interest.

\textbf{Data Availability.} All datasets used in this paper are available. 
Market-1501 (Unimodal ReID) is available at https://www.kaggle.com/datasets/pengcw1/market-1501. MSMT17 is available at https://www.kaggle.com/datasets/achrafabid7/msmt17.
SYSU-MM01 (VI-Reid) is available at https://github.com/wuancong/SYSU-MM01. RegDB (VI-Reid) is available at https://arxiv.org/abs/2001.04193. LLCM (VI-Reid) is available at https://github.com/ZYK100/LLCM. CUHK-PEDES (TI-Reid) is available at https://github.com/ShuangLI59/Person-Search-with-Natural-Language-Description. ICFG-PEDES (TI-Reid) is available at https://github.com/zifyloo/SSAN. RSTPReid (TI-Reid) is available at https://github.com/NjtechCVLab/RSTPReid-Dataset. PKU-Sketch (Sketch ReID) is available at https://www.pkuml.org/resources/pkusketchreid-dataset.html. CUFSF (Sketch ReID) is available at http://mmlab.ie.cuhk.edu.hk/archive/cufsf/. LReID and LReID-sync (NLOS ReID) are available at https://github.com/GWxuan/ReID3D. Wi-PER81 (NLOS ReID) is available at https://www.nature.com/articles/s41597-025-05804-0. Tri-CUHK-PEDES, Tri-ICFG-PEDES, and Tri-RSTPReid (Multimodal ReID) are available at https://github.com/ccq195/UNIReID. ORBench (Multimodal ReID) is available at https://github.com/zplusdragon/reid5o\_orbench.

\bibliography{sn-bibliography}

\end{document}